\documentclass{article}
\usepackage{amssymb}

\usepackage{graphics} 
\usepackage{epsfig} 
\usepackage{mathptmx} 
\usepackage{times} 
\usepackage{amsmath} 
\usepackage{amssymb}  
\usepackage{hyperref}
\usepackage{colortbl}
\usepackage{makecell}
\usepackage{multirow}
\usepackage{multicol}
\usepackage{color}
\usepackage{pifont}
\usepackage{wrapfig} 
 \usepackage{booktabs}
 \usepackage{enumitem}
 \usepackage{xspace}
\usepackage[usenames,dvipsnames,table]{xcolor}
\hypersetup{
    colorlinks=true,
    linkcolor=MidnightBlue,
    filecolor=magenta,      
    urlcolor=MidnightBlue,
    citecolor=MidnightBlue,
} 
\usepackage[font=small]{caption}

\usepackage[preprint]{corl_2025} 

\title{DEXOP: A Device for Robotic Transfer of Dexterous Human Manipulation}

\author{
\begin{tabular}{c}
\textbf{Hao-Shu Fang$^{1,2}$, Branden Romero$^{1,2*}$, Yichen Xie$^{3*}$, Arthur Hu$^{2*}$, Bo-Ruei Huang$^{2}$,} \\ \textbf{Juan Alvarez$^{2}$,   
Matthew Kim$^{2}$, Gabriel Margolis$^{1,2}$, Kavya Anbarasu$^{2}$,} \\ \textbf{Masayoshi Tomizuka$^{3}$, Edward Adelson$^{2}$ and Pulkit Agrawal$^{1,2}$} 
\end{tabular}\\
$^{1}$ Improbable AI Lab $^{2}$ Massachusetts Institute of Technology $^{3}$ UC Berkeley\\
\texttt{fhs@mit.edu, pulkitag@mit.edu} 
}

\newcommand{\pulkit}[1]{\textcolor{blue}{[Pulkit: #1]}}

\newcommand{\dexop}{\texttt{DEXOP}\xspace}
\newcommand{\dexops}{\texttt{DEXOP}-\texttt{7}\xspace}
\newcommand{\dexopn}{\texttt{DEXOP}-\texttt{9}\xspace}
\newcommand{\dexopt}{\texttt{DEXOP}-\texttt{12}\xspace}
\newcommand{\ours}{\xspace\textit{perioperation}\xspace}
\begin{document}
\maketitle

\renewcommand{\thefootnote}{}
\footnote{* indicates equal contribution; the order of these authors is interchangeable. Bo-Ruei Huang was a visiting student at MIT during his participation in this project.}
\setcounter{footnote}{0}

\vspace{-0.5in}
\begin{figure}[h]
  \centering
\includegraphics[width=1\linewidth]{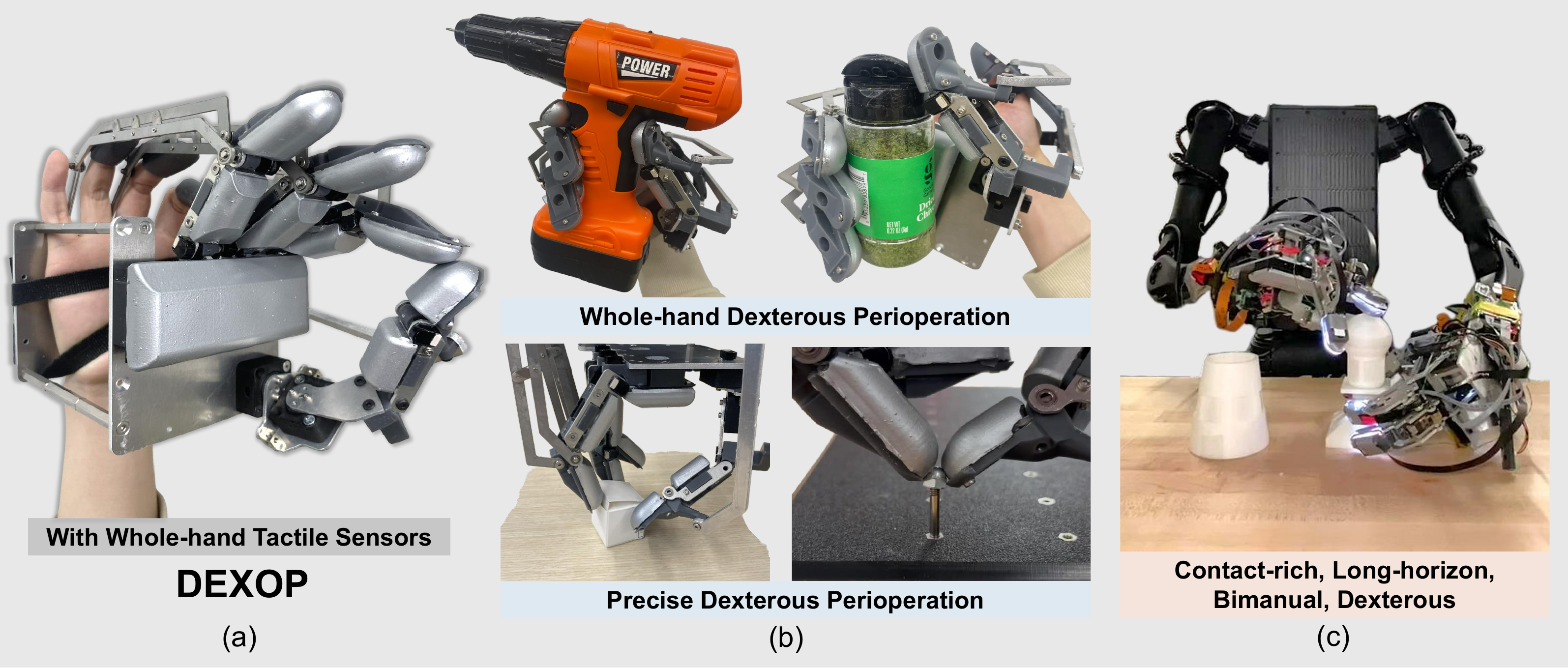}
\captionof{figure}{(a) \dexop is a passive exoskeleton that links human hand movements to passive robotic hand movements through mechanical linkages. (b) \dexop enables humans to collect task demonstrations of diverse and highly dexterous tasks. (c) Data collected by \dexop can be used to train policies that transfer to robots.} 
\label{fig:teaser}
  \vspace{-0.2in}
\end{figure}
\begin{abstract}
We introduce \textbf{\textit{perioperation}}, a paradigm for robotic data collection that sensorizes and records human manipulation while maximizing the transferability of the data to real robots. We implement this paradigm in \textbf{DEXOP}, a passive hand exoskeleton designed to maximize human ability to collect rich sensory (vision + tactile) data for diverse dexterous manipulation tasks in natural environments. 
\dexop mechanically connects human fingers to robot fingers, 
providing users with direct contact feedback (via proprioception) and mirrors the human hand pose to the passive robot hand to maximize the transfer of demonstrated skills to the robot. The force feedback and pose mirroring make task demonstrations more natural for humans compared to teleoperation, increasing both speed and accuracy. We evaluate \dexop across a range of dexterous, contact-rich tasks, demonstrating its ability to collect high-quality demonstration data at scale. Policies learned with \dexop data significantly improve task performance per unit time of data collection compared to teleoperation, making \dexop a powerful tool for advancing robot dexterity. Our project page is at \url{https://dex-op.github.io}.
\end{abstract}



\section{Introduction}
\label{sec:intro}

Dexterous manipulation is among the most challenging problems in robotics. Machine learning has significantly advanced robotic manipulation, but its reliance on large data poses a major bottleneck. 

Prominent ways of data collection include simulation, human activity videos, and teleoperation. Each has its pros and cons. Simulation based training~\cite{chen2022system, chen2023visual, qi2023general, andrychowicz2020learning, akkaya2019solving, zhu2019dexterous,torne2024reconciling,
ankile2024imitation,
park2024dexhub, huang2021generalization} allows large-scale, lower-cost experimentation with precise environmental control, but often requires extensive engineering to set up the simulation, and overcoming the sim-to-real gap~\cite{park2024position}. Human videos capture diverse environments and human expertise~\cite{antotsiou2018task, orbik2021human, qin2021dexmv, shaw2023videodex, mandikal2022dexvip, radosavovic2021state, wang2024dexcap}, but recovering fine-grained interaction and forces is an open challenge. 
Real-world robot data collection via teleoperation~\cite{qin2022one, yang2024ace, ding2024bunny, cheng2024open} provides robotic data that can directly be used for policy training without the challenges that come with sim2real or human video demonstrations. However, demonstrating dexterous manipulation is often unnatural for humans and remains costly to scale. More critically, providing humans with rich haptic feedback~\cite{sarakoglou2016hexotrac, zhang2025doglove} during teleoperation remains an open question. The absence of haptic feedback can substantially slow down the speed of teleoperation (and hence data collection) and limit demonstration of precise and fine manipulation tasks. 

In this paper, we advocate for a new approach to data collection called\textbf{\ours}: \textit{sensorizing human manipulation to capture rich multisensory data, including vision, proprioception, touch, and action, while maximizing the transferability of the demonstrated skills to robots}. The primary distinguishing factor from teleoperation is the focus on constructing devices that a human can wear and record data during their manipulation,  instead of remotely controlling a robot. 

\textbf{Challenges in sensorizing human manipulation:} In principle, we can sensorize human manipulation by having a person wear a glove that records hand pose and tactile data. However there are several pragmatic issues: (i) when the target robotic hand differs from the human hand—not only in morphology but also in physical properties such as compliance, friction, and skin softness—the collected data may not transfer well; (ii) most tactile gloves~\cite{SSundaram:2019:STAG, luo2024adaptive} use resistive sensors with lower spatial and force resolution than vision-based or Hall-effect based tactile sensors. It is debatable whether resistive tactile sensors provide information rich enough for a robot to reproduce human-level dexterity; and (iii) higher-fidelity tactile sensors (e.g., vision-based or Hall-effect) are usually rigid and bulky for humans to wear, impairing a user’s natural dexterity during manipulation. While these sensors can be used to sensorize the human fingertip~\cite{xing2025taccap, xu2025dexumi}, such information is insufficient for whole-hand manipulation.

Following\ours philosophy, we present \textbf{DEXOP}, a novel hand exoskeleton system designed to sensorize human manipulation while a user performs dexterous manipulation tasks in natural environments. Because we use high-resolution, but relatively bulky and rigid tactile sensors, \dexop installs these sensors on a passive robotic hand that is coupled through mechanical linkages to a passive exoskeleton (see Figure~\ref{fig:teaser}(a)) that a human can wear.  By moving the exoskeleton, the human user can actuate the passive hand to demonstrate diverse tasks (see Figure~\ref{fig:teaser}(b)). Three key design objectives guided the development of \dexop:
\begin{itemize}[wide, labelindent=0pt, labelwidth=0pt, itemsep=-0.0em]
    \item \textbf{Making Data Collection Natural:} \dexop makes data collection more natural and scalable by (i) maintaining high force transparency that enables users to perceive real-time and joint-level proprioceptive feedback through the robot hand, which addresses a primary limitation of current teleoperation methods, allowing faster (Section~\ref{sec:teleoperation}) and more precise (Section~\ref{sec:qual_res}) task demonstrations. (ii) Mapping the human hand pose to the passive robot hand's pose (i.e., kinematic coupling) through mechanical design eliminates the need for unintuitive visual hand pose correction during teleoperation, which often arises from inaccurate pose retargeting or kinematics mismatch between the human and the robot hand. This makes the task operation intuitive and thus easier to scale. (iii) Data collection without the full robot, which is more affordable and scalable to diverse environments.
    
    \item \textbf{High Transferability of Collected Data:} 
    Instead of using a tactile glove with limitations we discussed above, \dexop separates the human hand from the passive robotic hand, enabling the co-design of the passive and actual robotic hand to match their kinematic chain, shape, and sensors to maximize the transferability of collected data. Without this separation, it becomes difficult to replicate the same kinematic chain or sensors of the passive robotic hand on the human hand due to space constraints.
    For example, whole-hand tactile sensing is crucial for faithfully reproducing manipulation, because joint positions alone cannot guarantee the same interaction forces with objects. For instance, when grasping a mug, matching the hand pose without matching the contact forces can cause slip (too little force) or breakage (too much force). Since the object may contact the hand at multiple positions, densely capturing forces across the hand (i.e., whole-hand tactile sensing) allows us to preserve the action-relevant contact information needed for accurate and reliable replay on the robot. \dexop incorporates a whole-hand tactile sensing system (shown in Figure~\ref{fig:overview}(b)) that captures detailed force and contact information during interaction, similar to the EyeSight Hand~\cite{romero2024eyesight}. While \dexop presents whole-hand tactile sensing, we choose not to focus on using whole-hand sensing to recover the interaction forces in this paper. Instead, we focus on developing \dexop for demonstrating diverse tasks with the ability to capture whole-hand tactile sensing. We leave force recovery for future work, which will further improve the performance of policies trained with \dexop data.
    
    \item \textbf{Enhancing the diversity of accomplished tasks:} While \dexop's proprioceptive feedback can already enable users to do more tasks and be more efficient, we found that different mechanical enhancements can expand \dexop’s capabilities. (i) Adding fingernails allows \dexop to pick objects with shallow profile and manipulate small objects like an M2 screw cap. (ii) Abduction joints for the index/middle/ring fingers allow the fingertips to change the relative distance between each other, supporting better in-hand reorientation and manipulating objects of a larger size range. (iii) Finally, a padded palm can secure an object more firmly, making whole-hand manipulations—such as holding a seasoning bottle while opening the lid with the thumb—more stable. In \hyperref[supp:enhancements]{Supplementary Section S4}, we provide more details of these enhancements. These features have been explored in previous robotic hand designs~\cite{DenseTactMini2023, LaronSnePeretsSintov2024,BurgessAdelson2025, shaw2023leap, romero2024eyesight}, but not yet in data collection devices due to the challenges of adapting such designs while still allowing humans to manipulate objects effectively. Our approach of separating the human hand from the passive robotic hand makes their integration feasible. In Section~\ref{sec:device}, we summarize the comparison between \dexop and prior and concurrent data collection devices.
\end{itemize}
In this work, we developed three \dexop variants with different degrees of freedom (DoF), showing that the design framework of \dexop can easily be adapted to diverse robotic hands. We demonstrate the utility of \dexop across a variety of dexterous manipulation tasks, including drilling, lamp installation, box packaging, and bottle opening. Through comprehensive user studies, we show that \dexop provides superior control compared to traditional teleoperation systems and significantly improves data collection throughput. 
Our work lays the foundation for scalable, real-world robotic data collection and pushes the boundary of whole-hand and precise dexterous manipulation.

\section{Related Work}
\label{sec:citations}
\subsection{Teleoperation for Dexterous Manipulation}
Teleoperation is a widely used method for robotic data collection. Previous works used webcams, VR devices, or haptic gloves to teleoperate robotic systems~\cite{martinez2017towards, qin2022one, ding2024bunny, cheng2024open, yang2024ace, yin2025dexteritygen,park2024dexhub}. However, most teleoperation approaches either lack haptic feedback or provide limited feedback, for instance, using vibration cues, which are unintuitive. Other haptic feedback systems~\cite{sarakoglou2016hexotrac, zhang2025doglove} provide force feedback only at the fingertips and only the normal force, which can be limiting~\cite{luo2023team}. In contrast, \dexop provides haptic feedback with both normal and shear forces via the linkage system, and such feedback can optionally be expanded to each finger segment (implemented on \dexops, with details given in \hyperref[supp:kine]{Supplementary Section S2}).

\subsection{Data Collection Hardware for Robot Learning}
\label{sec:device}
Due to the growing demand for large-scale robotic data, hardware mechanisms for scalably collecting data have been of increasing interest in recent years. One line of work focuses on low-cost teleoperation systems~\cite{zhao2023learning, wu2023gello}, typically composed of a leader-follower setup where correspondence is achieved through joint mapping. Another line of work attempts to collect data without teleoperation by having a human directly control a passive robot gripper~\cite{fang2024airexo, chi2024universal, song2020grasping, shafiullah2023bringing} or by capturing human finger motions while performing object manipulation~\cite{wang2024dexcap, xing2025taccap}. Representative works like AirExo~\cite{fang2024airexo}, UMI~\cite{chi2024universal}, and DobbE~\cite{shafiullah2023bringing} use parallel-jaw grippers and capture only the position of the two-finger gripper. Our work advances this direction by (i) enabling manipulation that requires relative motion between fingers and objects, such as in-hand reorientation; (ii) improving manipulation efficiency compared to bi-manual parallel jaw gripper systems that require more steps for task completion; (iii) manipulation of small objects (e.g., screws, nuts) in constrained spaces which is hard to achieve with a bi-manual system, but can be achieved by a multi-fingered hand; (iv) additional degrees of freedom aslo enable manipulation of articulated objects such as spray bottles. Concurrently, several works~\cite{tao2025dexwild,xu2025dexumi,si2025exostart} explore passive exoskeletons for multi-fingered hands, but their device design does not emphasize fine/precise tasks nor whole-hand manipulation, resulting in the hand being largely used for basic open–close (flexion) motions that a simple gripper could also excel at. In contrast, \dexop unlocks a wider variety of tasks (whole-hand and fine manipulation), makes data collection superior (by recording force information), and more intuitive (via force transparency).

\subsection{Multi-Finger Hand Exoskeletons}
Hand exoskeletons have been extensively studied in both robotics and medical domains, primarily for rehabilitation, force augmentation, and haptic feedback~\cite{yun2017exo, kang2019exo, yun2016accurate, agarwal2015index, ferguson2020hand}. These systems feature motor systems to actively exert forces on human fingers, enhancing human motor control by amplifying movement and providing force feedback. In recent years, these systems have also been used for teleoperation and robot learning~\cite{sarakoglou2016hexotrac}. Our work, \dexop, took the opposite approach: instead of building a motorized exoskeleton to drive the human fingers, we developed a passive exoskeleton that allows humans to drive a passive robotic hand by moving their own fingers.

\begin{figure}[t!]
  \centering
\includegraphics[width=1\textwidth]{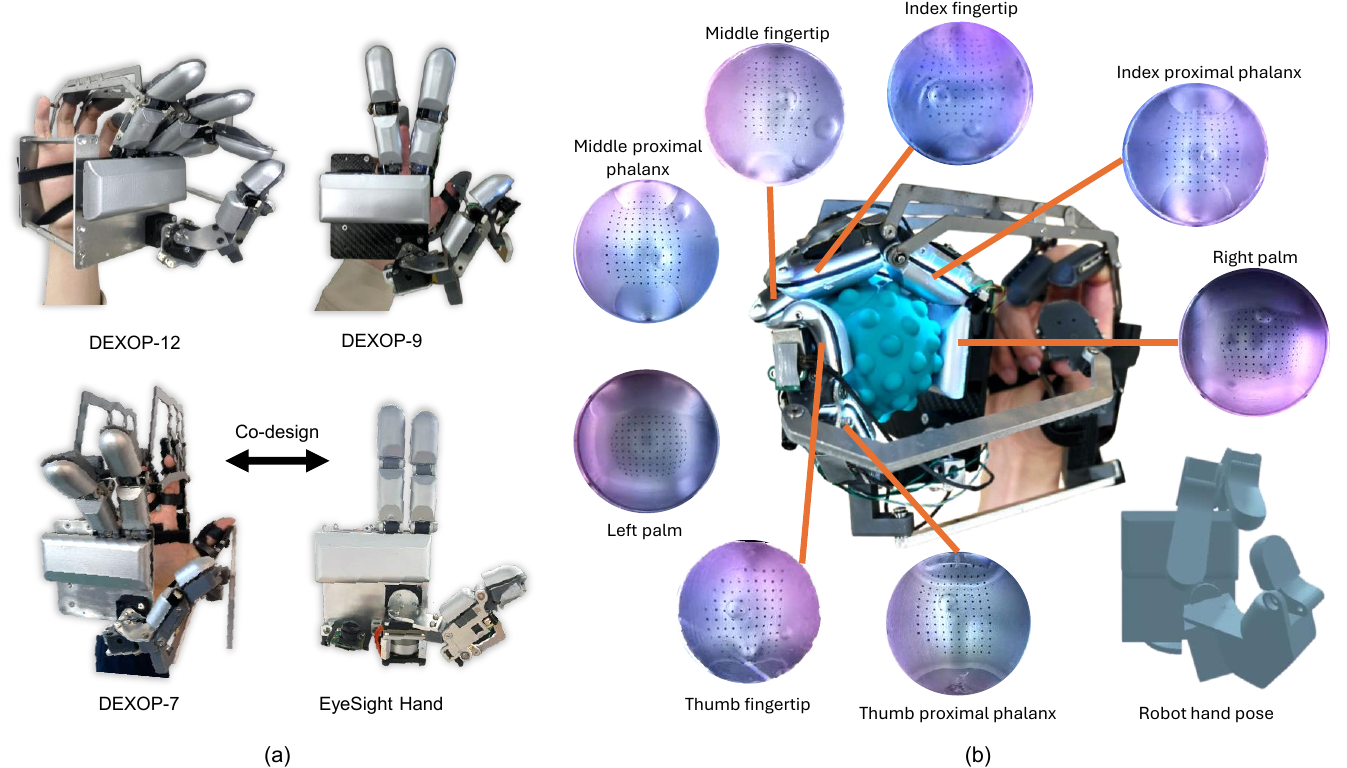}
  \caption{Hardware overview: (a) Variants of \dexop: \dexopt (4 fingers and 12 DOF), the most advanced \dexop; \dexopn without the ring finger (3 fingers and 9 DOF); and \dexops without abduction joints on index and middle fingers (3 fingers, 7 DOF, and co-designed with EyeSight hand). (b) Illustration of a human hand controlling the \dexopn to grasp a ball, along with the tactile sensor readings and robotic hand pose.}
  \label{fig:overview}
  \vspace{-0.1in}
\end{figure}

\section{Hardware Design}
\dexop consists of two main components: the passive robotic hand that interacts with objects and the wearable exoskeleton for the human hand. 
The passive robotic hand is mechanically linked to the exoskeleton through a linkage system. Forces applied by the human fingers to the exoskeleton are transmitted to the robotic hand, driving its motion. Conversely, interaction forces experienced by the passive robot hand are fed back to the human user through the same linkage and exoskeleton. 
\begin{wrapfigure}{r}{0.4\textwidth} 
  \centering
  \vspace{-0.1in}
\includegraphics[width=0.4\textwidth]{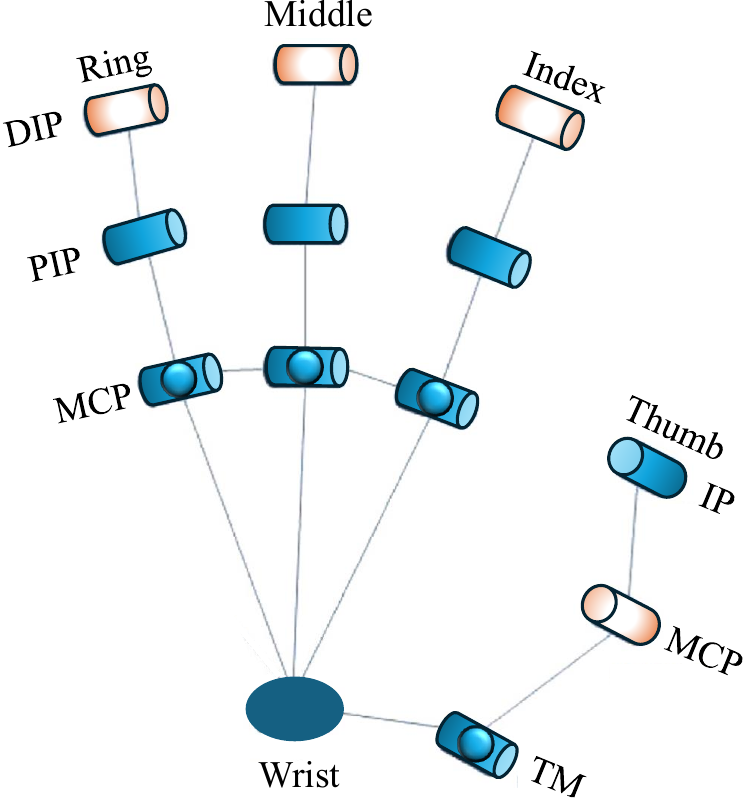}
  \caption{A depiction of joints in a human hand's kinematic chain. The blue joints are presented in \dexopt.}
  \label{fig:kinematics}
  \vspace{-0.5in}
\end{wrapfigure}

In this work, we present three \dexop variants: \dexopt with 4 fingers and 12 degrees of freedom, \dexopn with 3 fingers and 9 degrees of freedom, and \dexops with 3 fingers and 7 degrees of freedom. The first two are used to demonstrate the dexterous 
perioperation, while the third is \textit{co-designed} with a robotic hand~\cite{romero2024eyesight} for skill transfer to an actual robot. In this section, we detail the design of \texttt{DEXOP-12}—the most capable configuration—and provide descriptions of the other versions in \hyperref[supp:kine]{Supplementary Section S2}. Figure~\ref{fig:overview}(a) shows the different \dexop versions, and Figure~\ref{fig:overview}(b) shows an example of a human controlling \dexopn to grasp a ball and the captured data.

\subsection{Kinematics of the \dexop}
\label{sec:design}
For the \textit{passive robotic hand}, our design goal is to closely match the human finger's kinematic chain so that perioperation is more intuitive. \dexopt features 12 fully actuated degrees of freedom (DoF), with 3 DoF per finger. Figure~\ref{fig:kinematics} illustrates the kinematic chain of the human hand (colored in blue and orange) and shows which joints are implemented in \dexopt (colored in blue). 

The 12-DOF of \dexopt are chosen to maximize the diversity of tasks that can be performed: (i) Three fingers are required for single-hand in-hand manipulation, which is essential in constrained spaces and for tasks where the second hand is holding an object (e.g., holding a paper while the first hand uses the scissor). The fourth ring finger provides additional support for whole-hand manipulation tasks (e.g., stabilizing the seasoning bottle when opening the lid, see Figure~\ref{fig:qualitative}) (ii) The index, middle, and ring fingers each have a 2-DoF MCP joint and a PIP joint (refer to Figure~\ref{fig:kinematics} for these joint positions). The 2-DOF MCP joint is used to achieve abduction, which allows for changing the inter-finger distance and thereby increase/decrease the span of the hand for manipulating a wider range of object sizes and also aids in-hand object manipulation. (iii) The thumb has a 2-DoF TM joint and an IP joint. The TM joint enables \textit{flexion} motion~\cite{cooney1981kinesiology} of the thumb, which moves it to the opposite position of the index and middle fingers. This allows the hand to form antipodal grasps, aiding stable and precise manipulation. 

\dexopt doesn't feature a DIP joint on the index/middle/ring finger, nor the MCP joint on the thumb (shown in orange in Figure~\ref{fig:kinematics}). While including these joints could increase the diversity of tasks that can be performed, in our view the increase in complexity of the overall design of \dexop outweighs the potential increase in capabilities.

\begin{figure*}[t]
  \centering
  \includegraphics[width=0.95\textwidth]{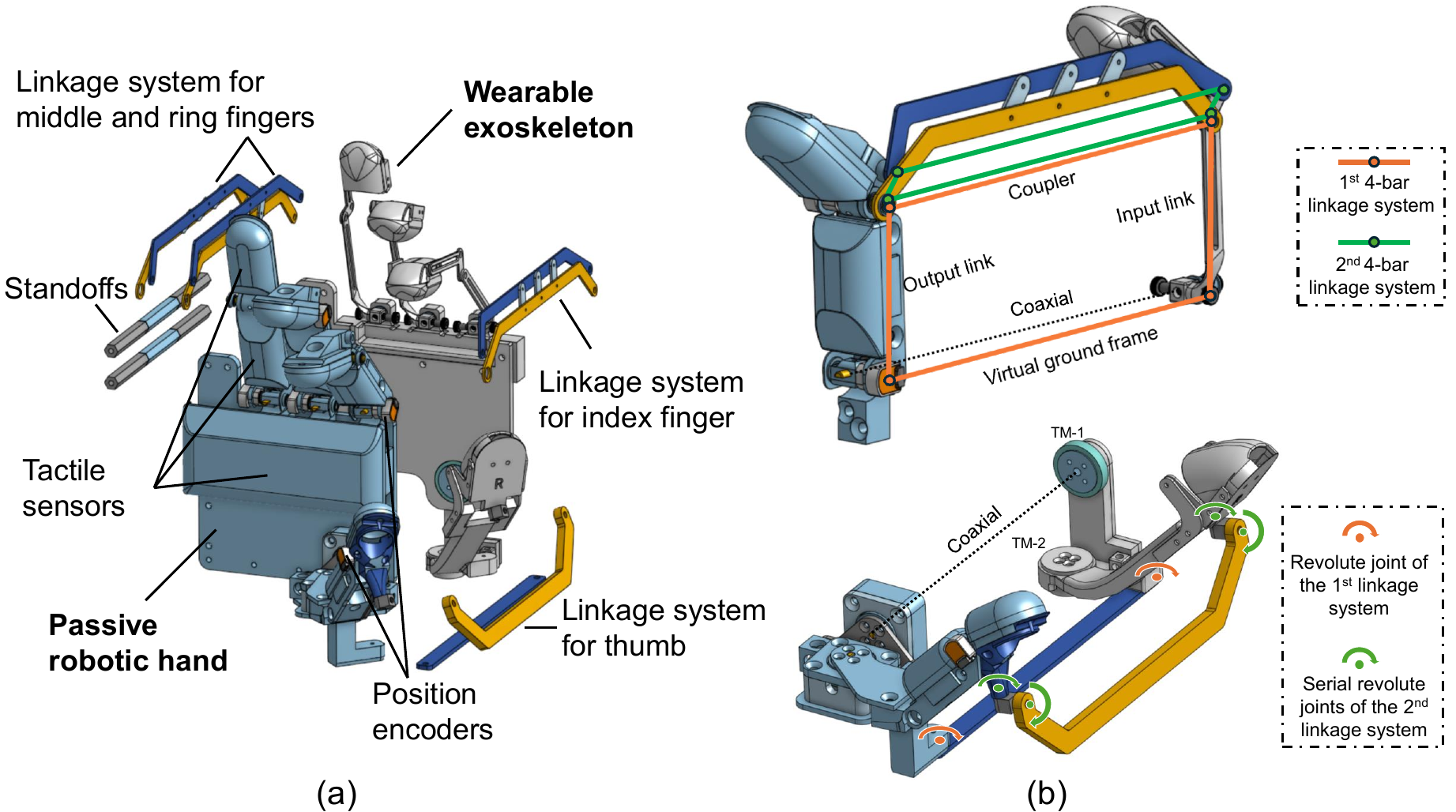}
  \caption{(a) Exploded view of the \dexopt system. (b) Top: Annotated view of the 4-bar linkages coupling the index, middle and ring fingers of the robot hand and exoskeleton. Bottom: Annotated view of the rotary linkage system coupling the thumb of the robot hand and exoskeleton. For a detailed explanation, please refer to text in Section~\ref{subsec:linkage-design}. 
  }
  \vspace{-0.2in}
  \label{fig:linkage}
\end{figure*}

For the \textit{wearable exoskeleton}, the human finger tip rests in a ``fingertip cot" that connects to the palm through exoskeleton segments (see the gray part in Figure~\ref{fig:linkage}(a)). If the fingertip cot is not connected to the palm (\textit{i.e.}, free floating), it is mechanically infeasible to design linkage systems that can couple unconstrained human finger motions to the passive robotic hand, as no one-to-one mapping exists. 

To facilitate simple control of the passive hand, the kinematic design of the exoskeleton matches the passive robotic hand’s kinematic chain, which enables motion transmission from exoskeleton to passive hand through 4-bar linkages (Section~\ref{subsec:linkage-design}). While in principle, the kinematic design can be different, doing so would complicate the linkage system without providing additional benefits. 

When implementing the wearable exoskeleton, two key challenges arise:

\begin{itemize}[wide, labelindent=0pt, labelwidth=0pt, itemsep=-0.0em]
\item 
For the index, middle, and ring fingers, collisions between the exoskeleton and the human fingers occur regardless of whether the exoskeleton is placed in front or behind the fingers, due to their close kinematic similarity and overlapping workspace during flexion (i.e., bending fingers towards the palm). To address this, the exoskeleton segments connecting the fingertip cot to the MCP joint are designed as thin sheet-metal links positioned alongside the fingers.

\item In the human hand, the two rotational axes of the thumb TM joint are positioned very closely. If the exoskeleton were designed with the same configuration, the limited spacing would cause the thumb exoskeleton to interfere with the user’s thumb, especially during flexion. To prevent this, we increase the perpendicular distance between the two TM joint axes, moving the abduction axis downward toward the wrist (see TM-1 and TM-2 at the bottom of Figure~\ref{fig:linkage}(b)). The thumb exoskeleton is then wrapped around the user’s thumb to avoid collisions during finger movement. 
\end{itemize}

\textbf{Co-design with actual robotic hand: }To maximize the transfer of demonstrated skills to the robot, the actual robotic hand needs to have the same kinematics chain as the passive robotic hand, thereby as the wearable exoskeleton. However, constructing an exoskeleton for the human hand that simply mimics the kinematic chain of an existing robotic hand can be uncomfortable for the user or be impossible to construct in a manner that allows human fingers to move naturally. Therefore, we co-design the passive and the real-robotic hand to ensure maximum transfer of collected data and the comfort for the human operator. In \hyperref[supp:codesign]{Supplementary Section S3}, we detail how we co-design \dexops by modifying the EyeSight Hand~\cite{romero2024eyesight}.

\subsection{Linkage Design}
\label{subsec:linkage-design}
The passive robotic hand is driven by the exoskeleton through a linkage system (see Figure~\ref{fig:linkage}(a)). Since the wearable exoskeleton (shown in gray) shares the same kinematics as the passive robotic hand (shown in light blue), we use multiple 4-bar linkages to control the fingers. In a 4-bar linkage system, a \textit{ground frame}—the link that remains stationary—is required. To realize this, we use standoffs to connect the base of the passive robotic hand and wearable exoskeleton, thereby fixing their relative distance and serving as a virtual ground frame. Reducing the standoff length can make \dexop less bulky, but if the distance is too short, it constrains human motion due to potential collisions with the passive robotic hand when human moves their fingers. Therefore, the standoff length is set to the minimum distance that avoids such collisions.

For the index, middle, and ring fingers, the linkage system is identical and illustrated on top of Figure~\ref{fig:linkage}(b). It consists of two 4-bar linkages that drive the two phalanges of each finger. In the first linkage, the fixed distance between the MCP joints of the exoskeleton and the passive robotic hand serves as the virtual ground frame. The exoskeleton's proximal phalanx functions as the input link, while the passive hand's proximal phalanx serves as the output. A curved link (shown in yellow) connects the two PIP joints of the wearable exoskeleton and the passive hand, acting as the coupler link. This 4-bar linkage system fixes the distance between the two PIP joints of the wearable exoskeleton and the passive hand, enabling the construction of a second 4-bar linkage system to actuate the distal phalanx. In this configuration, the coupler link of the first 4-bar linkage serves as the ground frame for the second, while the input and output links are the short axes of the distal phalanges on the exoskeleton and the passive hand. The coupler link (shown in dark blue) is also curved to provide additional clearance during movement 

For the thumb, the linkage system is shown at the bottom of Figure~\ref{fig:linkage}(b). The TM joint features two perpendicular axes, with the abduction joint of the exoskeleton and the passive robotic hand aligned coaxially. This allows a single coupler link (shown in dark blue) to drive the flexion and abduction axes of the TM joint, enabling control of two degrees of freedom. The IP joint, however, is not parallel to these axes. To control this additional degree of freedom, we introduce a second spatial 4-bar linkage. The coupler (shown in yellow) is connected to the distal phalanx via two serially arranged perpendicular joints. This spatial 4-bar linkage system can control the bending of the IP joint given any configuration of the TM joint.

\subsection{Tactile Sensor}
\label{sec:tactile-sensor}
As mentioned in Section~\ref{sec:intro}, replaying the joint position is insufficient for recreating the amount of the force the robot needs to exert. \textit{The missing information is the torque exerted by each joint}. In an exoskeleton setup, torque cannot be directly captured because it is generated by the human hand. 
A practical alternative is to capture force information at each phalanx along with joint positions, then compute joint torques using the Jacobian transpose method~\cite{chen2025dexforce}. Without whole-hand force sensing, however, this torque recovery problem becomes ill-posed: external forces may act at multiple contact points across the hand, and without observing these forces, the resulting torques cannot be uniquely or reliably inferred. 

In this work, we adopt the whole-hand tactile sensing design from the EyeSight Hand~\cite{romero2024eyesight}, equipping the passive robotic hand with GelSim(ple), a camera-based tactile sensor. The fingertips, palm, and proximal phalanges are all embedded with GelSim(ple) units. Each sensor uses one fisheye camera (or two in the case of the palm) with a 220° field of view for capturing deformations on the entire sensor's surface. Whole-hand sensing configuration significantly broadens the contact information in the collected data. For more details, we refer readers to the EyeSight Hand~\cite{romero2024eyesight}.

\subsection{In-the-wild Data Collection}

\dexop is a convenient tool for rapidly collecting dexterous manipulation data in natural environments. It can be integrated with an arm exoskeleton such as AirExo~\cite{fang2024airexo, fang2025airexo}, or use IMU/SLAM-based methods~\cite{shafiullah2023bringing, chi2024universal} to capture the global position of the hand, allowing a robot arm to replicate the motion of the hand. We also attach a fisheye camera to the base of the palm (close to the wrist) to capture visual observations. Data collection records global position (can be translation/rotation from SLAM or joint angles captured by arm exoskeleton), hand joint angles, tactile sensor images, in-hand camera views, and/or global scene images.

\section{Experiments}
\label{sec:exp}

In Sections~\ref{sec:capa} and~\ref{sec:teleoperation}, we evaluate the \dexop system by measuring its hardware characteristics and comparing its data collection efficiency against teleoperation. Both evaluations require a real robotic hand for meaningful comparison. Therefore, for these experimentss, we use the \dexops variant and its co-designed counterpart—the EyeSight Hand~\cite{romero2024eyesight}. Section~\ref{sec:qual_res} provides qualitative analysis of diverse tasks made possible by \dexop, where we use the \dexopn and \dexopt variants.

\subsection{Hardware Characteristics}
\label{sec:capa}
\begin{wraptable}{r}{0.5\columnwidth}
\centering
\small
\vspace{-0.1in}
\caption{Comparison of Force, Workspace, and Speed between \dexop System and Robotic Hand}
\label{tab:comparison}
\begin{tabular}{@{}cccc@{}}
\toprule
\makecell{\textbf{Metric}}            & \makecell{\textbf{Category}}    & \makecell{\textbf{\dexops}} & \makecell{\textbf{EyeSight}\\\textbf{Hand}} \\ \midrule
\multirow{3}{*}{\makecell{\textbf{Max Force}\\\textbf{(N)}}} & Thumb       & $\sim$70                  & 56                   \\
                                       & Index       & $\sim$60                  & 54                   \\
                                       & Middle      & $\sim$60                   & 54                         \\ \midrule
\multirow{7}{*}{\makecell{\textbf{Workspace}\\\textbf{(Degrees)}}} & MCP Joint  & 110                 & 120                                    \\
                                       & PIP Joint   & 105                 & 105                  \\
                                       & \makecell{TM joint\\(flexion)}   & 75                 & 75                  \\
                                       & \makecell{TM joint\\(abduction)}  & 90                 & 90                  \\
                                       & IP joint  & 65                 & 65                  \\\midrule
\multirow{7}{*}{\makecell{\textbf{Max Speed}\\\textbf{(rad/s)}}} & MCP Joint   & 35                 & 37                 \\
                                       & PIP Joint   & 15                 & 5                  \\
                                       & \makecell{TM joint\\(flexion)}  & 17                 & 32                  \\                                   & \makecell{TM joint\\(abduction)}  & 12                 & 35                  \\                                   & IP  & 9                 & 5                  \\ \bottomrule
\end{tabular}
\vspace{-0.3in}
\end{wraptable}
A good perioperation system should match the workspace, force, and speed of the robotic hand used to deploy the policy. Deficiencies in characteristics matching would indicate that the perioperation data cannot fully utilize the capabilities of the robotic system used for deploying policies, or are too hard for the robotic system to achieve. The performance of \dexops across several critical metrics (force output, workspace, and finger speed) and comparison against the EyeSight hand~\cite{romero2024eyesight} is summarized in Table~\ref{tab:comparison}, where we can see that \dexops hardware capability is comparable to the actual robotic hand.
The experimental procedure and comparison of each metric is presented below. 

\paragraph{Force Output} 
We wear \dexop and press the fingertip of the passive robotic hand on a force sensor while fixing the wrist on the table, to measure how much force can be applied at the fingertip. Our experiments provide evidence that \dexop effectively transmits forces from the human hand to the robotic hand. We measured a peak force of around \textbf{60 N} at the index and middle 
fingertips, and around \textbf{70 N} at the thumb fingertip, which is comparable to the force that the robotic hand can exert and similar to the maximum force that human fingertips can apply~\cite{an1986biomechanics}. 

\paragraph{Workspace Coverage}
When human subjects wear \dexop, its workspace closely mirrors the workspace of the robotic hand. The MCP joint on both systems allows approximately \textbf{110--120 degrees} of rotation, while the PIP joints reach up to \textbf{105 degrees}. The thumb’s motion on the \dexop system fully matches the robotic hand’s workspace across all three joints.

\paragraph{Finger Speed}
Finger speed was measured to assess how quickly the \dexop system responds to human input. \dexop's MCP joint reaches a maximum angular velocity of \textbf{35 rad/s}, slightly below the \textbf{37 rad/s} of the robotic hand. The PIP and IP joints (see Figure~\ref{fig:kinematics}) on \dexop achieve velocities of \textbf{15 rad/s} and \textbf{9 rad/s}, respectively, which is 2 to 3 times faster than those of the robotic hand. The TM joint on the \dexop system is comparatively slower, but it is worth noting that users must exert significant effort to reach these peak speeds with \dexop. In practice, such high speeds are rarely needed, as they may lead to unstable control in most manipulation tasks.

\begin{figure*}[t]
  \centering
  \includegraphics[width=1\textwidth]{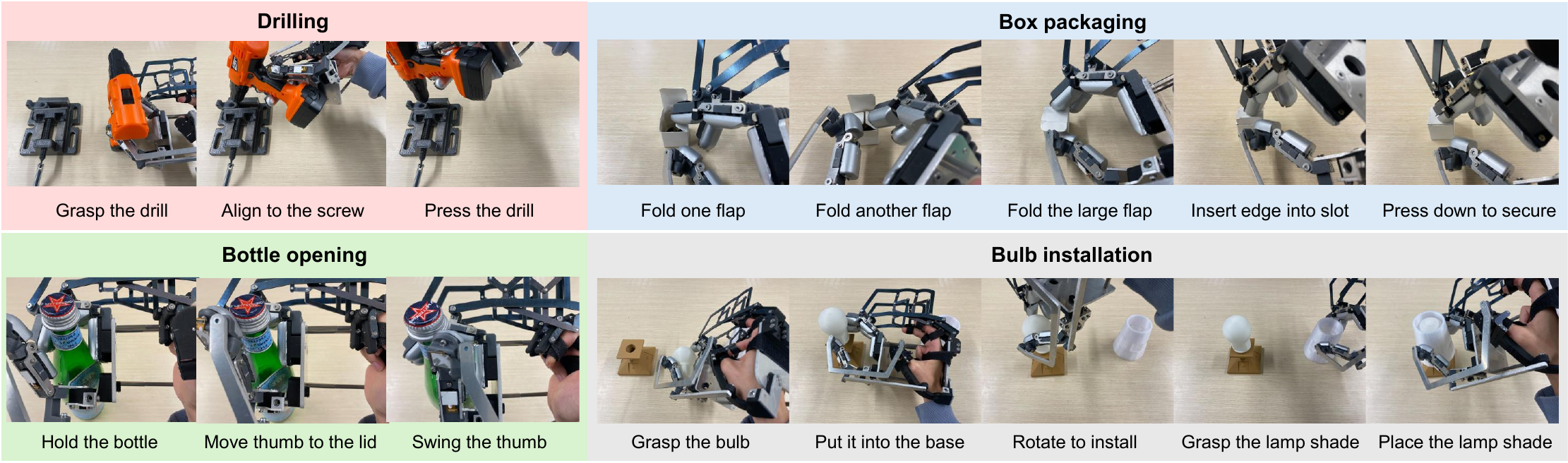}
  \caption{Illustration of evaluation tasks. \textbf{Drilling}: the user must pick up a drill standing upright on a table, then insert the drill bit into an M2 screw head and tighten it by actuating the drill. \textbf{Bottle opening}: with the bottle placed within the workspace of the hand, the user grasps the bottle and then uses the thumb to unscrew the cap. \textbf{Box Packaging:} the user approaches the an open box, and folds the side flaps before closing the top flap by folding the the securing flap into the box. \textbf{Bulb installation}: the task is composed of three parts, a lamp base, a light bulb, and a light shade. The user picks and screws the light bulb into the lamp base before placing the light shade over the entire assembly.}
  \label{fig:taskspec}
  \vspace{-0.1in}
\end{figure*}

\subsection{Comparison with Teleoperation}
\label{sec:teleoperation}

To evaluate the performance and usability of the \dexop system, we conducted a user study with four participants. Each participant performed four manipulation tasks, as illustrated in Figure~\ref{fig:taskspec}, using three different control schemes:

\begin{enumerate}[wide, labelindent=0pt, labelwidth=0pt]
    \item \textbf{\dexop System:} Participants controlled the passive robotic hand using the \dexops, enabling direct physical interaction through proprioceptive feedback.
    
    \item \textbf{Teleoperation:} A baseline teleoperation system composed of a UR3 robotic arm, a trakSTAR electromagnetic hand-tracking system, and an EyeSight hand was used. Participants manipulated the robotic hand via visual feedback, without haptic feedback. For more details on the tracking system, see~\cite{romero2024eyesight}.
    
    \item \textbf{Direct Human Performance:} As an upper-bound reference, participants completed the same tasks using their bare hands.
\end{enumerate}

\begin{figure*}[t]
  \centering
\includegraphics[width=0.92\textwidth]{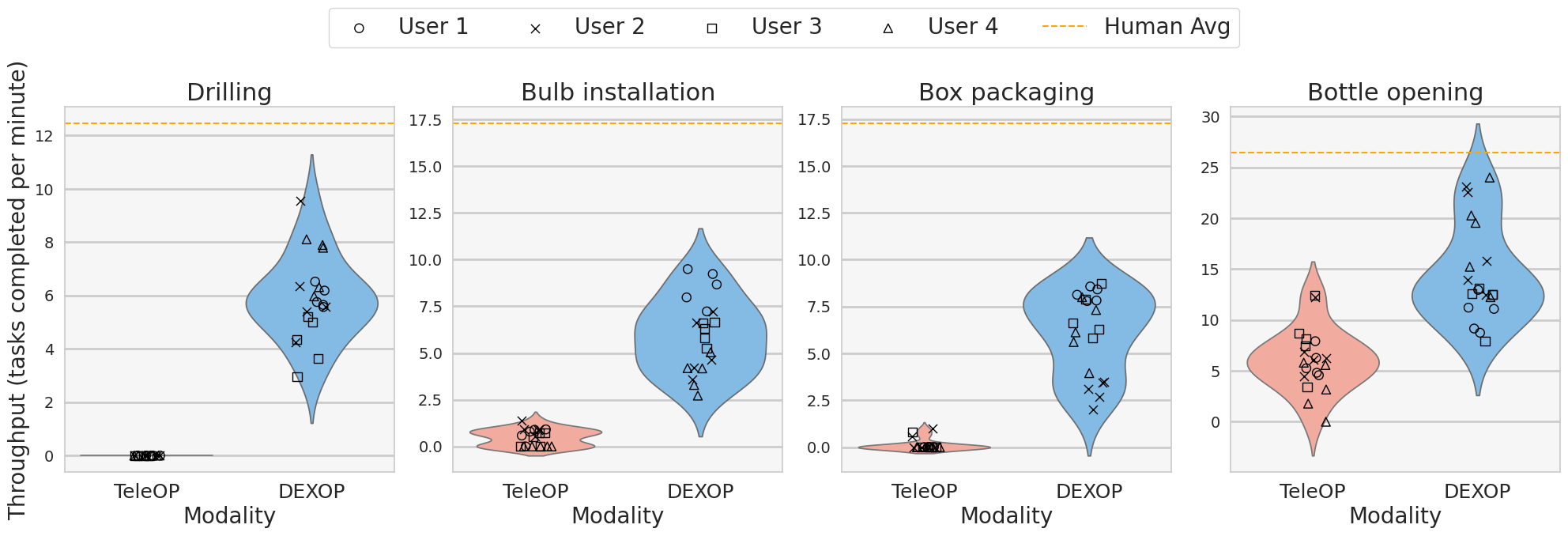}
  \caption{Comparison of task throughput of the drilling, bulb installation, box packaging and bottle opening tasks with TeleOP system, \dexop and human hand. \dexop achieves much higher throughput than TeleOP.}
  \label{fig:result}
  \vspace{-0.1in}
\end{figure*}

Each participant completed five trials for each task using each control scheme, resulting in a total of 240 trials across the four participants. Prior to task performance, participants were provided a brief task explanation and a 20-minute practice session. The primary performance metric was task throughput, calculated as the number of successful completions within one minute. Trials that exceeded three minutes were considered failures. The results of the user study are presented in Figure~\ref{fig:result}. Overall, DEXOP achieves much higher task throughput compared to the teleoperation system.

For the drilling task, participants encountered substantial difficulties using teleoperation. No participant successfully completed the task even once. The main failure mode was the difficulty in grasping the drill while keeping it operational. We posit that difficulty in teleoperation is caused by the limited visual feedback, especially for determining whether the index finger engaged the drill trigger. Aligning the drill with the small screw was also challenging. In contrast, with the \dexop system, participants completed the task an average of 6 times per minute. When using their own hands, participants achieved an average of 11 completions per minute.

For the bulb installation task, performance improved slightly under teleoperation: 15 out of 20 trials were successful, with an average completion time of 86 seconds. With the \dexop system, however, participants completed the task in just 11 seconds on average—eight times faster than with teleoperation. Using their own hands, participants completed the task in approximately 4 seconds.

The box packaging task was also highly challenging. Only 3 out of 20 trials were successful under teleoperation, with successful attempts averaging around 80 seconds. Most failures occurred when participants tried to fold the flaps—often pushing the box away in the process—or while attempting to insert the flap edge into the slot, which frequently caused the box to move or the flap to collapse. With the \dexop system, participants completed the task an average of 5 times per minute—7 times faster than teleoperation. Using their own hands, participants completed the task 16 times per minute.

The bottle opening task was found to be relatively easy by the participants. Teleoperation resulted in an average throughput of 5 times per minute. Using the \dexop system, participants achieved 12 completions per minute, 2.4$\times$ faster. When using their own hands, the average throughput reached 22 times per minute.

In conclusion, \dexop is much more efficient than teleoperation in accomplishing tasks that require proprioceptive feedback during manipulation, and is also a better data collection option on simple tasks such as bottle opening, where teleoperation can also accomplish effectively. Nevertheless, our experiments also suggest that this type of device has the potential for improvement to further approach human performance.

\begin{figure*}[t]
  \centering
\includegraphics[width=1.0\textwidth]{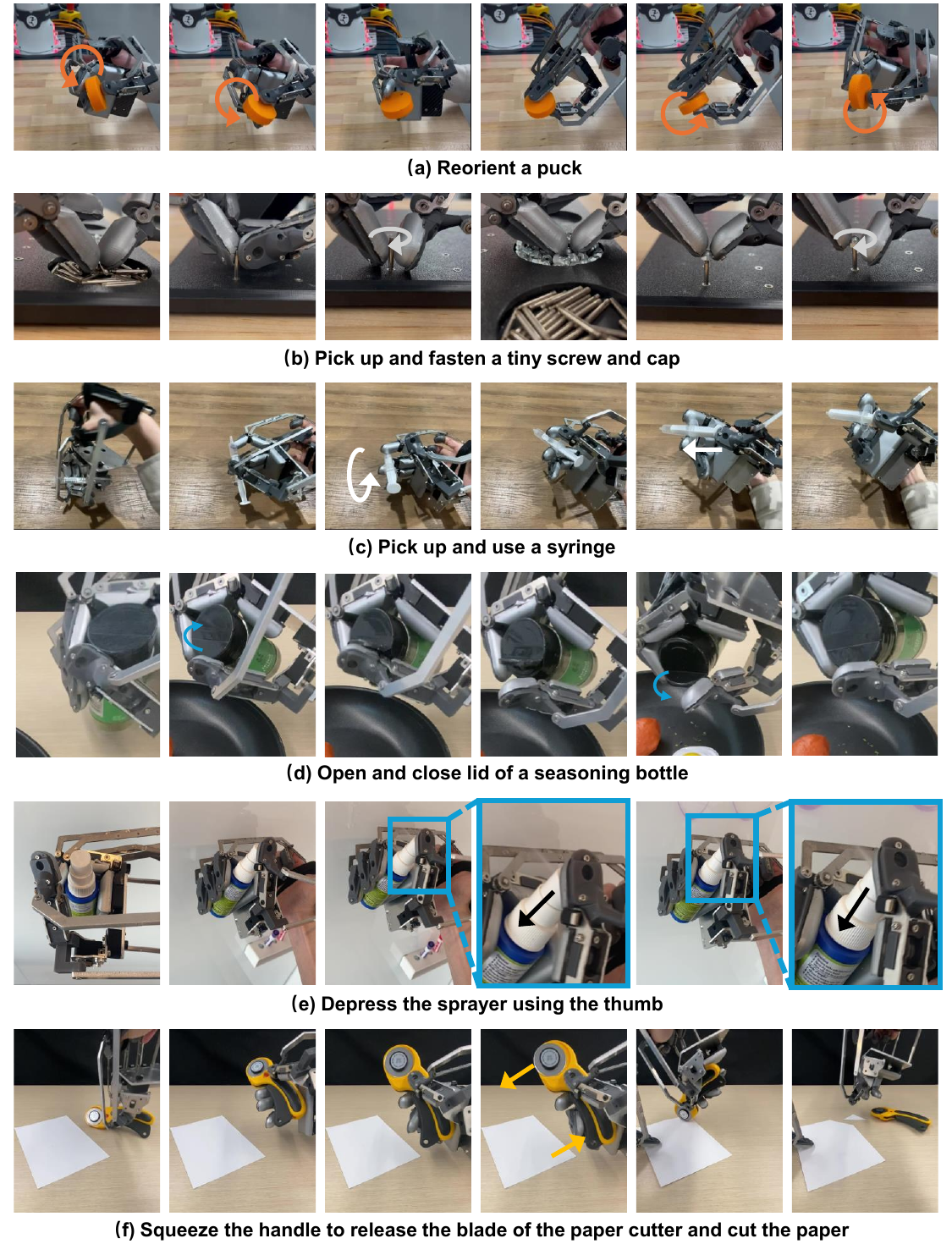}
  \caption{Various tasks illustrating the dexterity that \dexop can achieve. Arrows in the figure denote the motion of the object or its part being manipulated.}
  \label{fig:qualitative}
  \vspace{-0.1in}
\end{figure*}

\subsection{Dexterous Perioperation Ability}
\label{sec:qual_res}
The focus of our work is showcasing fine-grained dexterity with a focus on tasks that may not be achievable by parallel-jaw grippers. We categorize tasks into two types:

\textbf{(1) Precise finger manipulation}, which primarily involves fine finger motions. Examples include object reorientation, syringe operation, and tiny screw and cap manipulation—tasks that demand precision in fingertip control and pose adjustment of objects. 

\textbf{(2) Whole-hand manipulation}, which requires coordinated use of both the fingers and palm. It is pretty common during tool use~\cite{Napier1956,
Marzke1992}. Representative tasks include opening the lid of a seasoning bottle or depressing a sprayer with the thumb while constraining the bottle/sprayer with index/middle/ring fingers and the palm, and squeezing the handle of a paper cutter to release the blade and cut paper. These examples illustrate complex hand-environment interactions where force application and whole-hand contact are essential.

Qualitative demonstrations of both categories are shown in Figure~\ref{fig:qualitative}. For a more comprehensive view of \dexop’s capabilities, we invite readers to visit our website: \url{https://dex-op.github.io}. These results demonstrate \dexop's unique dexterous manipulation capabilities, which are not easily achievable via a teleoperated dexterous hand or parallel jaw grippers.

\section{Preliminary Policy Learning Experiments}
\label{sec:learning_experiments}
To evaluate the effectiveness of \dexop as a data collection tool for robot learning, we trained and evaluated policies learned from this data. We first describe the robot platform and how \dexop data is collected and aligned with the real robot. Then we describe the neural network architecture and tasks, and finally report the performance. As the corresponding robotic hands for \dexopn and \dexopt are still under development, to deploy the policy on the robot, we use \dexops and the co-designed version of EyeSight Hand~\cite{romero2024eyesight}. This ensures compatibility in both kinematics and sensing, allowing us to directly transfer data collected via perioperation to the real robot platform.

\subsection{Robot Platform Setup}
To evaluate policies trained with \dexop-collected data, we build two EyeSight Hands~\cite{romero2024eyesight} and mount them on a Unitree H1 humanoid robot (see Figure~\ref{fig:data_collection}(a)). Each EyeSight Hand has 3 fingers and 7 degrees of freedom. The Unitree H1 provides 4 degrees of freedom per arm, resulting in a combined 22-DoF system. During evaluation, the H1 is suspended from a gantry.
A fisheye camera is mounted on the base of the palm of each EyeSight Hand, near the wrist, to provide visual feedback.

\subsection{Data Collection Method}
\begin{figure*}[t]
  \centering
\includegraphics[width=1.0\textwidth]{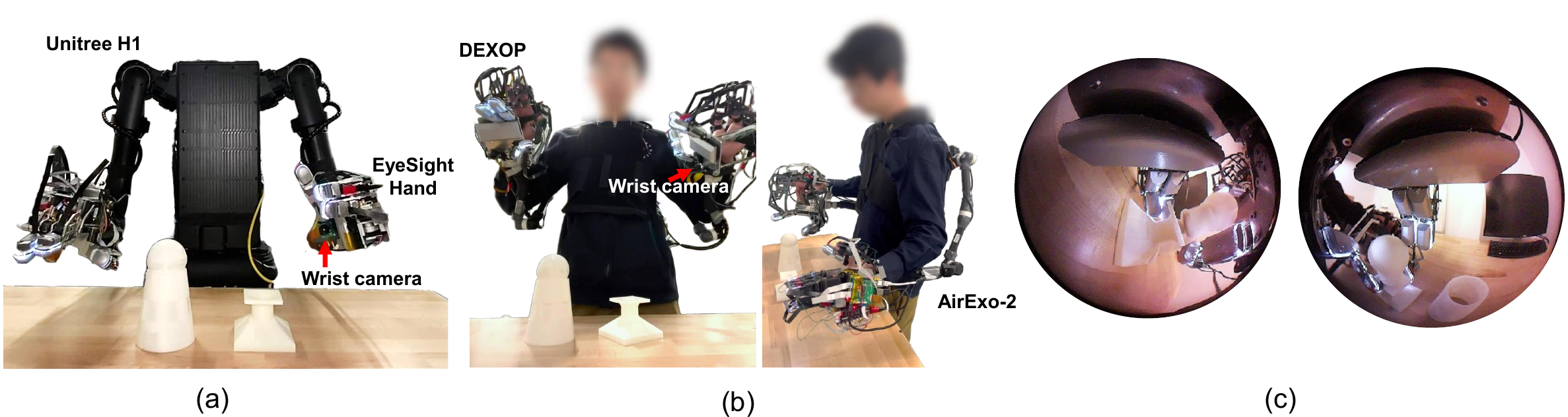}
  \caption{(a) A Unitree H1 with EyeSight hands used for policy evaluation. (b) The \dexop and AirExo setup, used for both teleop and \dexop data collection. (c) The views from the left and right wrist cameras mounted on the exoskeleton system.}
  \label{fig:data_collection}
  \vspace{-0.1in}
\end{figure*}

\dexop records finger joint positions, tactile images and wrist images. To capture global hand position, \dexop is mounted onto an in-house built whole-arm exoskeleton (AirExo-2~\cite{fang2025airexo}) customized to match the Unitree H1’s kinematics (see Figure~\ref{fig:data_collection}(b)). The AirExo-2 includes position encoders at each joint.

Whole-hand tactile sensing for both hands would require 4 Raspberry Pis and 4 ArduCam CamArray HATs (2 per hand). To make the electronics manageable, we enable tactile sensors only at the distal phalanx of each hand for this task, reducing the number of required Raspberry Pis and camera HATs to two. The tactile images from each hand are concatenated into a single super-image per hand. In total, we stream 22 joint angles, 2 wrist images (see Figure~\ref{fig:data_collection}(c)), and 2 tactile super-images at 20 Hz.

For comparison, we also collect teleoperation data. Since our \dexop+AirExo2 system tracks all the joints for the robot hand and upper body of H1, we directly map the joint angles from the exoskeleton to the robot, without the need to solve inverse kinematics. Note that this teleoperation setup does not introduce any errors due to kinematic differences and is already better than most teleoperation setups that need kinematic retargetting.

\subsection{Data Post-Processing}

Our \dexop system shares the same tactile sensors, kinematics, and visual configuration as the EyeSight Hand. Similarly, the customized AirExo-2 shares its kinematic structure with the Unitree H1. As a result, no further data processing is needed to close the embodiment gap; the data can be directly used to train the policy.

\begin{figure*}[t]
  \centering
\includegraphics[width=1.0\textwidth]{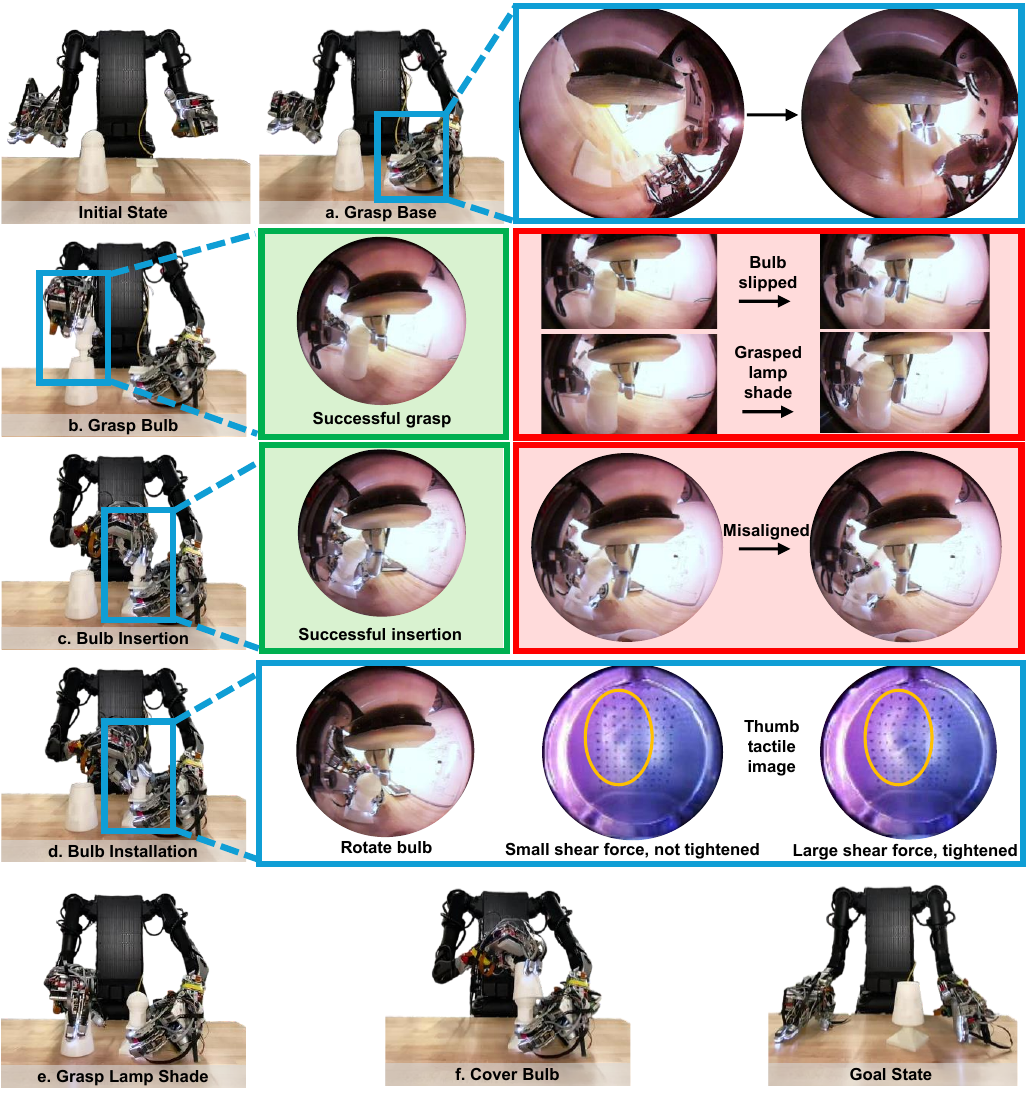}
  \caption{Illustration of our task for policy learning. Wrist camera images highlighted in green showcase successful actions for each stage, while images in red show different failure modes, such as misaligning the light bulb during insertion. Each of the six stages is shown in states after their respective actions have been performed successfully.}
  \label{fig:policy_illu}
  \vspace{-0.1in}
\end{figure*}

\subsection{Task Setup}
\label{sec:bulb-install}
To evaluate the transferability of \dexop-collected data to real-world robot execution, we evaluated performance on a complex bimanual manipulation task, \textbf{bulb installation},  modified from the lamp task from FurnitureBench~\cite{heo2023furniturebench}. This task is long-horizon, contact-rich, and requires both fine dexterous control and coordinated bimanual action. It integrates multiple low-level skills under partial tactile and visual observability. We decompose the task into six key stages, as shown in Figure~\ref{fig:policy_illu}, each targeting a specific manipulation capability:

\begin{enumerate}[wide, labelwidth=!, labelindent=0pt, label=(\Alph*)]
    \item \textbf{Grasp Base:} The robot must reliably grasp the lamp base from a resting position. This stage evaluates the robustness of basic grasping skills under positional variations in the environment.

    \item \textbf{Grasp Bulb:} The lamp is a spherical object, which makes grasping more sensitive to misalignment. The robot must orient its fingers precisely along the object’s diameter to avoid slipping, making this step a test of fine grasp alignment.

    \item \textbf{Bulb Insertion:} This is a high-precision phase requiring the lamp to be aligned and inserted into the base socket. Success depends on both accurate motion planning and closed-loop correction based on real-time vision and tactile feedback. If the insertion is not accurate initially, the robot must detect misalignment and adjust accordingly.

    \item \textbf{Bulb Installation:} Once inserted, the robot must rotate the lamp to screw it into the base. This step requires coordinated multi-finger manipulation, along with perception of subtle torque or force changes to determine whether the bulb is securely installed. The robot must use tactile sensing to infer task state and trigger a transition to the next stage.

    \item \textbf{Grasp Lamp Shade:} This stage revisits basic grasping, this time for the lamp cover. Success depends on recognizing object geometry and executing a reliable power grasp.

    \item \textbf{Cover Bulb:} Finally, the robot must place the cover over the assembled lamp, requiring coordinated motion between both arms and fine alignment to avoid collision or tilt. This evaluates bimanual coordination and spatial reasoning with a large object.
\end{enumerate}

This complete sequence tests a wide spectrum of dexterous capabilities, including grasp robustness, finger coordination, fine alignment, tactile-informed decision making, and bimanual coordination, making it a strong benchmark for learning from \dexop-collected demonstrations.

\subsection{Learning Setup}
\label{sec:learning-setup}
To validate that data collected using \dexop can be effectively used for training dexterous manipulation policies, we trained policies using behavior cloning. 
We use the Action Chunking Transformer (ACT)~\cite{zhao2023learning} as our policy architecture. The inputs to the policy include:
\begin{itemize}[wide, labelindent=0pt, labelwidth=0pt]
    \item Two wrist camera images (one per hand), providing visual context.
    \item Two tactile super-images (one per hand), each contains three concatenated tactile images from the thumb/index/middle distal phalanges. To emphasize contact changes, we use the delta image between the current and initial tactile readings.
    \item The current joint state, including hand and arm configurations. This input is used as a query following the original ACT implementation.
\end{itemize}

The network outputs:
\begin{itemize}[wide, labelindent=0pt, labelwidth=0pt]
    \item \textbf{Delta joint positions} for robot arms. This improves robustness to fabrication variability and calibration noise from the exoskeleton hardware, since relative motion is more transferable than absolute position in such systems.
    \item \textbf{Absolute joint positions} for the hands. Since our hand hardware is precisely machined and co-designed with \dexop, absolute joint commands generalize well without significant error accumulation.
\end{itemize}

To improve generalization and robustness during training, we apply several augmentations:
\begin{itemize}[wide, labelindent=0pt, labelwidth=0pt]
    \item \textbf{Color jittering} (brightness and hue randomized within ±0.1) is applied to wrist camera images to account for lighting variation.
    \item \textbf{Joint noise} of up to ±10 degrees is injected into the arm states with a 10\% chance, simulating variability in human arm posture or assembly imprecision.
    \item \textbf{Dropout} of 0.3 is applied to the wrist image stream, following prior work~\cite{romero2024eyesight}, to prevent over-reliance on vision and encourage integration of tactile information.
\end{itemize}

\begin{figure*}[t]
  \centering
\includegraphics[width=1.0\textwidth]{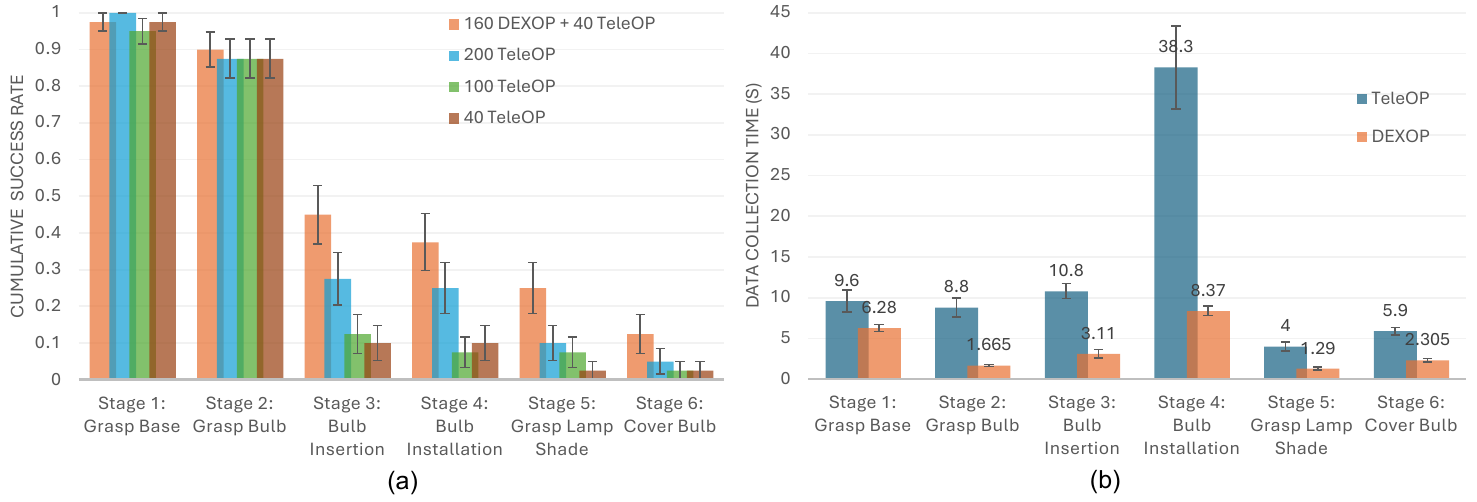}
  \caption{(a) Success rates of policies trained on mixed \dexop and TeleOP data, broken down by task step. Note that these are cumulative success rates, where performance in earlier stages affects the outcomes of subsequent stages. (b) Speed comparison of data collection for \dexop and TeleOP, averaged time for each trajectory by task step. Error bars denote the standard error of the mean.}
  \label{fig:results}
  \vspace{-0.1in}
\end{figure*}

\subsection{Results of \dexop and TeleOP}
We train different policies on different data sources collected by \dexop and teleop and compare their performance. 
While in principle, we can directly use the data collected by \dexop to train the policy, we observed accumulated joint errors in the arm exoskeleton data due to in-house fabrication variability—e.g., errors in cutting carbon fiber tubes and assembling components. During data collection, human operators may also introduce variability by bending their backs, which causes additional misalignment between the exoskeleton and the robot because we fix the robot in an upright position in this paper. While this issue could be addressed by mounting the arm exoskeleton on a stationary base frame~\cite{fang2025airexo}, or using IMU sensors to track torso orientation and compensate for it on the robot platform, doing so is beyond the scope of this work.
\if
there are two challenges in practice:
\begin{itemize}[wide, labelindent=0pt, labelwidth=0pt]
    \item With captured joint positions and whole-hand tactile sensing, we can recover joint torques for robot action using the Jacobian transpose method~\cite{studywolf2013jacobian, asada1986robotics, chen2025dexforce}. However, it requires accurate force estimates. For our in-house built vision-based tactile sensors, extracting force estimates from images  (force calibration) remains complex \pulkit{but this was the whole motivation for tactile sensing in the first half of the paper and now we are saying we can't do it.}.
    \item We observed accumulated joint errors in the arm exoskeleton data due to in-house fabrication variability—e.g., errors in cutting carbon fiber tubes and assembling components. During data collection, human operators may also introduce variability by bending their backs, which causes additional misalignment between the exoskeleton and the robot because we fix the robot in an upright position in this paper. While this issue could be addressed by mounting the arm exoskeleton on a stationary base frame~\cite{fang2025airexo}, or using IMU sensors to track torso orientation and compensate for it on the robot platform, doing so is beyond the scope of this work.
\end{itemize}
\fi

To mitigate this challenge, we augment the \dexop data with a small amount of teleoperated robot demo and co-train the policy on a mix of \dexop and teleop data. The reason is that during teleoperation, the human can observe the motion of the robot and compensate for the errors of the arm exoskeleton, which wouldn't happen in \dexop data collection.  We trained four policies under different conditions:  
\begin{itemize}[wide, labelindent=0pt, labelwidth=0pt]
    \item \textbf{\dexop + TeleOP (160 + 40 demos):} 160 demonstrations collected using \dexop combined with 40 demonstrations collected via teleoperation.  
    \item \textbf{TeleOP (200 demos):} 200 demonstrations collected via teleoperation, matching the total number of demonstrations in the first condition.  
    \item \textbf{TeleOP (100 demos):} 100 demonstrations collected via teleoperation, matching the total data collection time of the first condition, given \dexop’s approximately $2.67\times$ faster data collection rate on this task.  
    \item \textbf{TeleOP (40 demos):} 40 demonstrations collected via teleoperation, serving as a baseline without any \dexop data compared to the first condition.  
\end{itemize}

Training is conducted on an NVIDIA H100 GPU with a batch size of 110, and each model is trained for 800 epochs. For the model using a combination of DEXOP and TeleOP data, training is performed on the mixed dataset for 500 epochs, followed by 300 epochs on the TeleOP subset to help calibrate errors in the arm exoskeleton assembly.

For each policy, we repeat evaluations 40 times with small perturbations ($\pm$2 cm translation, $\pm$5$^\circ$ rotation) on object location. The results are given in Figure~\ref{fig:results}(a). In Table~\ref{tab:success}, we report the total data collection time for each condition and their respective normalized cumulative success rate over all the 6 steps, $\frac{\sum_{i=1}^{6} S_i}{6}$.

\begin{wraptable}{r}{0.5\columnwidth}
    \centering
    \vspace{-0.1in}
    \small
    \caption{Data collection time and normalized cumulative success.}
    \vspace{-0.1in}
    \begin{tabular}{lcc}
        \toprule
        \textbf{Method} & \textbf{Time (mins)} & \textbf{Success} \\
        \midrule
        160 \dexop + 40 TeleOP & 139.3 & 0.513$_{\pm0.032}$\\
        200 TeleOP             & 283.3 & 0.425$_{\pm0.032}$ \\
        100 TeleOP             & 141.7 & 0.355$_{\pm0.031}$ \\
        40 TeleOP              & 56.7  & 0.350$_{\pm0.031}$ \\
        \bottomrule
    \end{tabular}
    \vspace{-0.2in} 
    \label{tab:success}
\end{wraptable}

The 160 \dexop + 40 teleop policy achieves the highest success rates across all stages, particularly excelling in challenging steps of \textit{bulb insertion} and \textit{grasp lamp cover}. In contrast, the 40 teleop policy shows the poorest performance, with rapid degradation in later stages. Although the data collection time for 100 teleop and 160 \dexop + 40 teleop is the same, the performance of 100 teleop is considerably worse. This result highlights that \dexop provides a more efficient way to collect data for training robust policies.

We surprisingly observe that the 200 teleop policy is worse than 160 \dexop + 40 teleop, despite teleoperation data generally being considered high quality. Specifically, two stages contribute most to this performance gap: the bulb insertion and grasp lamp cover steps. For bulb insertion, we hypothesize that exoskeleton-based data introduces greater variation (less precise) in object positioning during collection, because operators tend to be more relaxed when doing perioperation than teleoperation. 

For lamp cover grasping, we observe that the 200 teleop policy often fails by continuously rotating the bulb without transitioning to grasping the lamp cover. To better understand this behavior, we analyze the time spent in each stage during data collection and report the results in Figure~\ref{fig:results}(b). The results show that, during teleoperation, operators spend substantially more time (38 VS 6 seconds) than \dexop in the screw bulb into base stage due to the lack of haptic feedback. Without shear force sensing, it is difficult for operators to judge if the bulb is fully tightened, often leading to over-rotation. \textit{This overemphasis introduces a bias in the dataset, causing the policy to prioritize rotation behaviors instead of transitioning to subsequent steps}, leading to frequent failures in the grasp lamp shade stage.

These findings suggest that teleoperation not only reduces data collection efficiency but also introduces systematic biases due to limited sensory feedback to operators, ultimately degrading policy performance. On the contrary, \dexop provides rich force feedback, leading not only to faster demonstrations, but also to \textit{cleaner, less biased data that better reflect successful human strategies}, ultimately improving policy generalization and performance in downstream tasks.


\section{Discussion}
\label{sec:discussion}
\paragraph{Perioperation for large-scale data collection.} Perioperation systems are sensitive to manufacturing defects in the exoskeleton and to sensor calibration. Mismatches between the target robot platform and the exoskeleton can degrade performance to the extent that DEXOP data alone may not be sufficient to deploy a learned policy. In this paper, we compensated for such errors by incorporating teleoperation data. One way to address this issue is to improve the calibration of the exoskeleton. Other exciting avenues include: (i) treating \dexop as a data-collection tool for pre-training a foundation model, with small errors compensated by real robot data obtained through either real-world reinforcement learning or teleoperation; and (ii) developing more powerful manipulation models that are robust to minor errors in robot action labels. While perioperation systems require additional design effort, we believe that investing in their development and then distributing them offers unmatched scalability compared to teleoperation.

\paragraph{Dexterity needs careful design.} When designing \dexop, we found it relatively easy to enable a multi-finger hand to grasp large objects, but far from trivial to grasp small objects (with one dimension smaller than 5mm) or manipulate articulated objects using its redundant degrees of freedom. Releasing the potential of dexterity requires careful design, as mentioned in \hyperref[supp:enhancements]{Supplementary Section S4}. Without such attention to detail, we found that perioperation with a multi-finger hand can be inferior to using a simple parallel-jaw gripper, especially when grasping small objects. \dexop incorporates many of these design considerations, though there is still room for improvement in the future such as adding a DIP joint and optimizing the side and back finger shapes.

\paragraph{Limitations and future work.} Several limitations remain in our current paper. First, estimating joint torques from tactile and kinematic data still requires sensor calibration and real-time inference. Second, the current EyeSight hand does not have the degrees of freedom that the human hand does to allow us to use \dexop data to perform very dexterous tasks like in-hand reorientation. Third, the current \dexop system provides proprioceptive feedback but still misses the tactile feedback to humans. Fourth, we have only scratched the surface of what learning from \ours data can achieve; future work will explore more learning methods, like how to combine tactile and vision that play different roles in different stages of manipulation, and how to enable precise force control during manipulation.

\paragraph{Broader vision.} As robot hands become more capable, the lack of high-quality data remains a major bottleneck. We believe that systems like \dexop fill a crucial missing layer between raw human demonstrations and robotic generalization. By making it easier to capture rich, tactile, contact-driven data at scale, we hope this work will accelerate the co-evolution of better data, better hardware, and ultimately, more dexterous robots.

\acknowledgments{ We thank the members of the Improbable AI lab for the helpful discussions and feedback on the paper. We thank Lirui Wang for helping with the human study of perioperation versus teleoperation, Sameen Ahmad and Xavier Sanchez for helping with assembling the EyeSight hand and the \dexop. This project benefits from the AAU program of Arducam. The authors sincerely thank their support on multiple cameras data collection system, especially Ajax. This
 research was partly supported by Toyota Research Institute and Magna Inc. We acknowledge
 support from ONR MURI under grant number N00014-22-1-2740. The views and conclusions contained in this document
 are those of the authors and should not be interpreted as
 representing the official policies, either expressed or implied,
 of the Army Research Office or the U.S. Government. 
}

\subsection*{Author contributions}
\small
\textbf{Project formulation:} Hao-Shu Fang, Pulkit Agrawal; 
\textbf{Paper writing:} Hao-Shu Fang, Pulkit Agrawal; 
\textbf{\dexop design:} Hao-Shu Fang, Arthur Hu, Juan Alvarez; 
\textbf{\dexop building:} Hao-Shu Fang, Arthur Hu; 
\textbf{EyeSight hand co-design:} Branden Romero, Hao-Shu Fang; 
\textbf{EyeSight hand building:} Branden Romero; 
\textbf{AirExo2 building:} Matthew Kim; 
\textbf{\dexop driver:} Bo-Ruei Huang; 
\textbf{Teleoperation infrastructure:} Bo-Ruei Huang, Hao-Shu Fang, Gabriel Margolis; 
\textbf{Data collection:} Hao-Shu Fang, Kavya Anbarasu, Matthew Kim; 
\textbf{Learning and deployment:} Hao-Shu Fang, Yichen Xie; 
\textbf{Advice and support for hardware manufacturing:} Edward Adelson, Pulkit Agrawal\\

\bibliography{example}  

\begin{thebibliography}{56}
\providecommand{\natexlab}[1]{#1}
\providecommand{\url}[1]{\texttt{#1}}
\expandafter\ifx\csname urlstyle\endcsname\relax
  \providecommand{\doi}[1]{doi: #1}\else
  \providecommand{\doi}{doi: \begingroup \urlstyle{rm}\Url}\fi

\bibitem[Chen et~al.(2022)Chen, Xu, and Agrawal]{chen2022system}
T.~Chen, J.~Xu, and P.~Agrawal.
\newblock A system for general in-hand object re-orientation.
\newblock In \emph{Conference on Robot Learning}, pages 297--307. PMLR, 2022.

\bibitem[Chen et~al.(2023)Chen, Tippur, Wu, Kumar, Adelson, and Agrawal]{chen2023visual}
T.~Chen, M.~Tippur, S.~Wu, V.~Kumar, E.~Adelson, and P.~Agrawal.
\newblock Visual dexterity: In-hand reorientation of novel and complex object shapes.
\newblock \emph{Science Robotics}, 8\penalty0 (84):\penalty0 eadc9244, 2023.

\bibitem[Qi et~al.(2023)Qi, Yi, Suresh, Lambeta, Ma, Calandra, and Malik]{qi2023general}
H.~Qi, B.~Yi, S.~Suresh, M.~Lambeta, Y.~Ma, R.~Calandra, and J.~Malik.
\newblock General in-hand object rotation with vision and touch.
\newblock In \emph{Conference on Robot Learning}, pages 2549--2564. PMLR, 2023.

\bibitem[Andrychowicz et~al.(2020)Andrychowicz, Baker, Chociej, Jozefowicz, McGrew, Pachocki, Petron, Plappert, Powell, Ray, et~al.]{andrychowicz2020learning}
O.~M. Andrychowicz, B.~Baker, M.~Chociej, R.~Jozefowicz, B.~McGrew, J.~Pachocki, A.~Petron, M.~Plappert, G.~Powell, A.~Ray, et~al.
\newblock Learning dexterous in-hand manipulation.
\newblock \emph{The International Journal of Robotics Research}, 39\penalty0 (1):\penalty0 3--20, 2020.

\bibitem[Akkaya et~al.(2019)Akkaya, Andrychowicz, Chociej, Litwin, McGrew, Petron, Paino, Plappert, Powell, Ribas, et~al.]{akkaya2019solving}
I.~Akkaya, M.~Andrychowicz, M.~Chociej, M.~Litwin, B.~McGrew, A.~Petron, A.~Paino, M.~Plappert, G.~Powell, R.~Ribas, et~al.
\newblock Solving rubik's cube with a robot hand.
\newblock \emph{arXiv preprint arXiv:1910.07113}, 2019.

\bibitem[Zhu et~al.(2019)Zhu, Gupta, Rajeswaran, Levine, and Kumar]{zhu2019dexterous}
H.~Zhu, A.~Gupta, A.~Rajeswaran, S.~Levine, and V.~Kumar.
\newblock Dexterous manipulation with deep reinforcement learning: Efficient, general, and low-cost.
\newblock In \emph{2019 International Conference on Robotics and Automation (ICRA)}, pages 3651--3657. IEEE, 2019.

\bibitem[Torne et~al.(2024)Torne, Simeonov, Li, Chan, Chen, Gupta, and Agrawal]{torne2024reconciling}
M.~Torne, A.~Simeonov, Z.~Li, A.~Chan, T.~Chen, A.~Gupta, and P.~Agrawal.
\newblock Reconciling reality through simulation: A real-to-sim-to-real approach for robust manipulation.
\newblock \emph{arXiv preprint arXiv:2403.03949}, 2024.

\bibitem[Ankile et~al.()Ankile, Simeonov, Shenfeld, Villasevil, and Agrawal]{ankile2024imitation}
L.~L. Ankile, A.~Simeonov, I.~Shenfeld, M.~T. Villasevil, and P.~Agrawal.
\newblock From imitation to refinement--residual rl for precise visual assembly.
\newblock In \emph{CoRL 2024 Workshop on Mastering Robot Manipulation in a World of Abundant Data}.

\bibitem[Park et~al.(2024)Park, Bhatia, Ankile, and Agrawal]{park2024dexhub}
Y.~Park, J.~S. Bhatia, L.~Ankile, and P.~Agrawal.
\newblock Dexhub and dart: Towards internet scale robot data collection.
\newblock \emph{arXiv preprint arXiv:2411.02214}, 2024.

\bibitem[Huang et~al.(2021)Huang, Mordatch, Abbeel, and Pathak]{huang2021generalization}
W.~Huang, I.~Mordatch, P.~Abbeel, and D.~Pathak.
\newblock Generalization in dexterous manipulation via geometry-aware multi-task learning.
\newblock \emph{arXiv preprint arXiv:2111.03062}, 2021.

\bibitem[Park et~al.()Park, Margolis, and Agrawal]{park2024position}
Y.~Park, G.~B. Margolis, and P.~Agrawal.
\newblock Position: Automatic environment shaping is the next frontier in rl.
\newblock In \emph{Forty-first International Conference on Machine Learning}.

\bibitem[Antotsiou et~al.(2018)Antotsiou, Garcia-Hernando, and Kim]{antotsiou2018task}
D.~Antotsiou, G.~Garcia-Hernando, and T.-K. Kim.
\newblock Task-oriented hand motion retargeting for dexterous manipulation imitation.
\newblock In \emph{Proceedings of the European Conference on Computer Vision (ECCV) Workshops}, pages 0--0, 2018.

\bibitem[Orbik et~al.(2021)Orbik, Li, and Lee]{orbik2021human}
J.~Orbik, S.~Li, and D.~Lee.
\newblock Human hand motion retargeting for dexterous robotic hand.
\newblock In \emph{2021 18th International Conference on Ubiquitous Robots (UR)}, pages 264--270. IEEE, 2021.

\bibitem[Qin et~al.(2021)Qin, Wu, Liu, Jiang, Yang, Fu, and Wang]{qin2021dexmv}
Y.~Qin, Y.-H. Wu, S.~Liu, H.~Jiang, R.~Yang, Y.~Fu, and X.~Wang.
\newblock Dexmv: Imitation learning for dexterous manipulation from human videos.
\newblock \emph{arXiv preprint arXiv:2108.05877}, 2021.

\bibitem[Shaw et~al.(2023)Shaw, Bahl, and Pathak]{shaw2023videodex}
K.~Shaw, S.~Bahl, and D.~Pathak.
\newblock Videodex: Learning dexterity from internet videos.
\newblock In \emph{Conference on Robot Learning}, pages 654--665. PMLR, 2023.

\bibitem[Mandikal and Grauman(2022)]{mandikal2022dexvip}
P.~Mandikal and K.~Grauman.
\newblock Dexvip: Learning dexterous grasping with human hand pose priors from video.
\newblock In \emph{Conference on Robot Learning}, pages 651--661. PMLR, 2022.

\bibitem[Radosavovic et~al.(2021)Radosavovic, Wang, Pinto, and Malik]{radosavovic2021state}
I.~Radosavovic, X.~Wang, L.~Pinto, and J.~Malik.
\newblock State-only imitation learning for dexterous manipulation.
\newblock In \emph{2021 IEEE/RSJ International Conference on Intelligent Robots and Systems (IROS)}, pages 7865--7871. IEEE, 2021.

\bibitem[Wang et~al.(2024)Wang, Shi, Wang, Zhang, Fei-Fei, and Liu]{wang2024dexcap}
C.~Wang, H.~Shi, W.~Wang, R.~Zhang, L.~Fei-Fei, and C.~K. Liu.
\newblock Dexcap: Scalable and portable mocap data collection system for dexterous manipulation.
\newblock \emph{arXiv preprint arXiv:2403.07788}, 2024.

\bibitem[Qin et~al.(2022)Qin, Su, and Wang]{qin2022one}
Y.~Qin, H.~Su, and X.~Wang.
\newblock From one hand to multiple hands: Imitation learning for dexterous manipulation from single-camera teleoperation.
\newblock \emph{arXiv preprint arXiv:2204.12490}, 2022.

\bibitem[Yang et~al.(2024)Yang, Liu, Qin, Ding, Li, Cheng, Yang, Yi, and Wang]{yang2024ace}
S.~Yang, M.~Liu, Y.~Qin, R.~Ding, J.~Li, X.~Cheng, R.~Yang, S.~Yi, and X.~Wang.
\newblock Ace: A cross-platform visual-exoskeletons system for low-cost dexterous teleoperation.
\newblock \emph{arXiv preprint arXiv:2408.11805}, 2024.

\bibitem[Ding et~al.(2024)Ding, Qin, Zhu, Jia, Yang, Yang, Qi, and Wang]{ding2024bunny}
R.~Ding, Y.~Qin, J.~Zhu, C.~Jia, S.~Yang, R.~Yang, X.~Qi, and X.~Wang.
\newblock Bunny-visionpro: Real-time bimanual dexterous teleoperation for imitation learning.
\newblock \emph{arXiv preprint arXiv:2407.03162}, 2024.

\bibitem[Cheng et~al.(2024)Cheng, Li, Yang, Yang, and Wang]{cheng2024open}
X.~Cheng, J.~Li, S.~Yang, G.~Yang, and X.~Wang.
\newblock Open-television: teleoperation with immersive active visual feedback.
\newblock \emph{arXiv preprint arXiv:2407.01512}, 2024.

\bibitem[Sarakoglou et~al.(2016)Sarakoglou, Brygo, Mazzanti, Hernandez, Caldwell, and Tsagarakis]{sarakoglou2016hexotrac}
I.~Sarakoglou, A.~Brygo, D.~Mazzanti, N.~G. Hernandez, D.~G. Caldwell, and N.~G. Tsagarakis.
\newblock Hexotrac: A highly under-actuated hand exoskeleton for finger tracking and force feedback.
\newblock In \emph{2016 IEEE/RSJ International Conference on Intelligent Robots and Systems (IROS)}, pages 1033--1040. IEEE, 2016.

\bibitem[Zhang et~al.(2025)Zhang, Hu, Yuan, and Xu]{zhang2025doglove}
H.~Zhang, S.~Hu, Z.~Yuan, and H.~Xu.
\newblock Doglove: Dexterous manipulation with a low-cost open-source haptic force feedback glove.
\newblock \emph{arXiv preprint arXiv:2502.07730}, 2025.

\bibitem[Sundaram et~al.(2019)Sundaram, Kellnhofer, Li, Zhu, Torralba, and Matusik]{SSundaram:2019:STAG}
S.~Sundaram, P.~Kellnhofer, Y.~Li, J.-Y. Zhu, A.~Torralba, and W.~Matusik.
\newblock Learning the signatures of the human grasp using a scalable tactile glove.
\newblock \emph{Nature}, 569\penalty0 (7758), 2019.
\newblock \doi{10.1038/s41586-019-1234-z}.

\bibitem[Luo et~al.(2024)Luo, Liu, Lee, DelPreto, Wu, Foshey, Rus, Palacios, Li, Torralba, et~al.]{luo2024adaptive}
Y.~Luo, C.~Liu, Y.~J. Lee, J.~DelPreto, K.~Wu, M.~Foshey, D.~Rus, T.~Palacios, Y.~Li, A.~Torralba, et~al.
\newblock Adaptive tactile interaction transfer via digitally embroidered smart gloves.
\newblock \emph{Nature communications}, 15\penalty0 (1):\penalty0 868, 2024.

\bibitem[Xing et~al.(2025)Xing, Li, Wei, Ren, Tu, Lin, Schumann, Zheng, and Cutkosky]{xing2025taccap}
C.~Xing, H.~Li, Y.-L. Wei, T.-A. Ren, T.~Tu, Y.~Lin, E.~Schumann, W.-S. Zheng, and M.~R. Cutkosky.
\newblock Taccap: A wearable fbg-based tactile sensor for seamless human-to-robot skill transfer.
\newblock \emph{arXiv e-prints}, pages arXiv--2503, 2025.

\bibitem[Xu et~al.(2025)Xu, Zhang, Hou, Xu, Fan, Veloso, and Song]{xu2025dexumi}
M.~Xu, H.~Zhang, Y.~Hou, Z.~Xu, L.~Fan, M.~Veloso, and S.~Song.
\newblock Dexumi: Using human hand as the universal manipulation interface for dexterous manipulation.
\newblock \emph{arXiv preprint arXiv:2505.21864}, 2025.

\bibitem[Romero et~al.(2024)Romero, Fang, Agrawal, and Adelson]{romero2024eyesight}
B.~Romero, H.-S. Fang, P.~Agrawal, and E.~Adelson.
\newblock Eyesight hand: Design of a fully-actuated dexterous robot hand with integrated vision-based tactile sensors and compliant actuation.
\newblock \emph{arXiv preprint arXiv:2408.06265}, 2024.

\bibitem[Do et~al.(2023)Do, Dhawan, Kitzmann, and III]{DenseTactMini2023}
W.~K. Do, A.~K. Dhawan, M.~Kitzmann, and M.~K. III.
\newblock Densetact-mini: An optical tactile sensor for grasping multi-scale objects from flat surfaces.
\newblock 2023.
\newblock arXiv preprint arXiv:2309.08860.

\bibitem[Laron et~al.(2024)Laron, Sne, Perets, and Sintov]{LaronSnePeretsSintov2024}
A.~Laron, E.~Sne, Y.~Perets, and A.~Sintov.
\newblock Print-n-grip: A disposable, compliant, scalable and one-shot 3d-printed multi-fingered robotic hand.
\newblock 2024.
\newblock arXiv preprint arXiv:2401.16463.

\bibitem[Burgess and Adelson(2025)]{BurgessAdelson2025}
M.~Burgess and E.~H. Adelson.
\newblock Grasp everything (get): 1-dof, 3-fingered gripper with tactile sensing for robust grasping.
\newblock 2025.
\newblock arXiv preprint arXiv:2505.09771.

\bibitem[Shaw et~al.(2023)Shaw, Agarwal, and Pathak]{shaw2023leap}
K.~Shaw, A.~Agarwal, and D.~Pathak.
\newblock Leap hand: Low-cost, efficient, and anthropomorphic hand for robot learning.
\newblock \emph{arXiv preprint arXiv:2309.06440}, 2023.

\bibitem[Martinez-Hernandez et~al.(2017)Martinez-Hernandez, Szollosy, Boorman, Kerdegari, and Prescott]{martinez2017towards}
U.~Martinez-Hernandez, M.~Szollosy, L.~W. Boorman, H.~Kerdegari, and T.~J. Prescott.
\newblock Towards a wearable interface for immersive telepresence in robotics.
\newblock In \emph{Interactivity, Game Creation, Design, Learning, and Innovation: 5th International Conference, ArtsIT 2016, and First International Conference, DLI 2016, Esbjerg, Denmark, May 2--3, 2016, Proceedings 5}, pages 65--73. Springer, 2017.

\bibitem[Yin et~al.(2025)Yin, Wang, Pineda, Hogan, Bodduluri, Sharma, Lancaster, Prasad, Kalakrishnan, Malik, et~al.]{yin2025dexteritygen}
Z.-H. Yin, C.~Wang, L.~Pineda, F.~Hogan, K.~Bodduluri, A.~Sharma, P.~Lancaster, I.~Prasad, M.~Kalakrishnan, J.~Malik, et~al.
\newblock Dexteritygen: Foundation controller for unprecedented dexterity.
\newblock \emph{arXiv preprint arXiv:2502.04307}, 2025.

\bibitem[Luo et~al.(2023)Luo, Wang, Keil, Nguyen, Mayne, Alt, Schwarm, Mendoza, Pad{\i}r, and Whitney]{luo2023team}
R.~Luo, C.~Wang, C.~Keil, D.~Nguyen, H.~Mayne, S.~Alt, E.~Schwarm, E.~Mendoza, T.~Pad{\i}r, and J.~P. Whitney.
\newblock Team northeastern's approach to ana xprize avatar final testing: A holistic approach to telepresence and lessons learned.
\newblock In \emph{2023 IEEE/RSJ International Conference on Intelligent Robots and Systems (IROS)}, pages 7054--7060. IEEE, 2023.

\bibitem[Zhao et~al.(2023)Zhao, Kumar, Levine, and Finn]{zhao2023learning}
T.~Z. Zhao, V.~Kumar, S.~Levine, and C.~Finn.
\newblock Learning fine-grained bimanual manipulation with low-cost hardware.
\newblock \emph{arXiv preprint arXiv:2304.13705}, 2023.

\bibitem[Wu et~al.(2023)Wu, Shentu, Yi, Lin, and Abbeel]{wu2023gello}
P.~Wu, Y.~Shentu, Z.~Yi, X.~Lin, and P.~Abbeel.
\newblock Gello: A general, low-cost, and intuitive teleoperation framework for robot manipulators.
\newblock \emph{arXiv preprint arXiv:2309.13037}, 2023.

\bibitem[Fang et~al.(2024)Fang, Fang, Wang, Ren, Chen, Zhang, Wang, and Lu]{fang2024airexo}
H.~Fang, H.-S. Fang, Y.~Wang, J.~Ren, J.~Chen, R.~Zhang, W.~Wang, and C.~Lu.
\newblock Airexo: Low-cost exoskeletons for learning whole-arm manipulation in the wild.
\newblock In \emph{2024 IEEE International Conference on Robotics and Automation (ICRA)}, pages 15031--15038. IEEE, 2024.

\bibitem[Chi et~al.(2024)Chi, Xu, Pan, Cousineau, Burchfiel, Feng, Tedrake, and Song]{chi2024universal}
C.~Chi, Z.~Xu, C.~Pan, E.~Cousineau, B.~Burchfiel, S.~Feng, R.~Tedrake, and S.~Song.
\newblock Universal manipulation interface: In-the-wild robot teaching without in-the-wild robots.
\newblock \emph{arXiv preprint arXiv:2402.10329}, 2024.

\bibitem[Song et~al.(2020)Song, Zeng, Lee, and Funkhouser]{song2020grasping}
S.~Song, A.~Zeng, J.~Lee, and T.~Funkhouser.
\newblock Grasping in the wild: Learning 6dof closed-loop grasping from low-cost demonstrations.
\newblock \emph{IEEE Robotics and Automation Letters}, 5\penalty0 (3):\penalty0 4978--4985, 2020.

\bibitem[Shafiullah et~al.(2023)Shafiullah, Rai, Etukuru, Liu, Misra, Chintala, and Pinto]{shafiullah2023bringing}
N.~M.~M. Shafiullah, A.~Rai, H.~Etukuru, Y.~Liu, I.~Misra, S.~Chintala, and L.~Pinto.
\newblock On bringing robots home.
\newblock \emph{arXiv preprint arXiv:2311.16098}, 2023.

\bibitem[Tao et~al.(2025)Tao, Srirama, Liu, Shaw, and Pathak]{tao2025dexwild}
T.~Tao, M.~K. Srirama, J.~J. Liu, K.~Shaw, and D.~Pathak.
\newblock Dexwild: Dexterous human interactions for in-the-wild robot policies.
\newblock \emph{arXiv preprint arXiv:2505.07813}, 2025.

\bibitem[Si et~al.(2025)Si, Chen, Karagozler, Bronars, Hutchinson, Lampe, Gileadi, Howell, Saliceti, Barczyk, et~al.]{si2025exostart}
Z.~Si, J.~E. Chen, M.~E. Karagozler, A.~Bronars, J.~Hutchinson, T.~Lampe, N.~Gileadi, T.~Howell, S.~Saliceti, L.~Barczyk, et~al.
\newblock Exostart: Efficient learning for dexterous manipulation with sensorized exoskeleton demonstrations.
\newblock \emph{arXiv preprint arXiv:2506.11775}, 2025.

\bibitem[Yun et~al.(2017)Yun, Kang, and Cho]{yun2017exo}
S.-S. Yun, B.~B. Kang, and K.-J. Cho.
\newblock Exo-glove pm: An easily customizable modularized pneumatic assistive glove.
\newblock \emph{IEEE Robotics and Automation Letters}, 2\penalty0 (3):\penalty0 1725--1732, 2017.

\bibitem[Kang et~al.(2019)Kang, Choi, Lee, and Cho]{kang2019exo}
B.~B. Kang, H.~Choi, H.~Lee, and K.-J. Cho.
\newblock Exo-glove poly ii: A polymer-based soft wearable robot for the hand with a tendon-driven actuation system.
\newblock \emph{Soft robotics}, 6\penalty0 (2):\penalty0 214--227, 2019.

\bibitem[Yun et~al.(2016)Yun, Agarwal, Fox, Madden, and Deshpande]{yun2016accurate}
Y.~Yun, P.~Agarwal, J.~Fox, K.~E. Madden, and A.~D. Deshpande.
\newblock Accurate torque control of finger joints with ut hand exoskeleton through bowden cable sea.
\newblock In \emph{2016 IEEE/RSJ International Conference on Intelligent Robots and Systems (IROS)}, pages 390--397. IEEE, 2016.

\bibitem[Agarwal et~al.(2015)Agarwal, Fox, Yun, O’Malley, and Deshpande]{agarwal2015index}
P.~Agarwal, J.~Fox, Y.~Yun, M.~K. O’Malley, and A.~D. Deshpande.
\newblock An index finger exoskeleton with series elastic actuation for rehabilitation: Design, control and performance characterization.
\newblock \emph{The International Journal of Robotics Research}, 34\penalty0 (14):\penalty0 1747--1772, 2015.

\bibitem[Ferguson et~al.(2020)Ferguson, Shen, and Rosen]{ferguson2020hand}
P.~W. Ferguson, Y.~Shen, and J.~Rosen.
\newblock Hand exoskeleton systems—overview.
\newblock \emph{Wearable Robotics}, pages 149--175, 2020.

\bibitem[Cooney et~al.(1981)Cooney, Lucca, Chao, and Linscheid]{cooney1981kinesiology}
W.~P. Cooney, M.~J. Lucca, E.~Y.~S. Chao, and R.~L. Linscheid.
\newblock The kinesiology of the thumb trapeziometacarpal joint.
\newblock \emph{The Journal of Bone and Joint Surgery. American Volume}, 63\penalty0 (9):\penalty0 1371--1381, 1981.
\newblock \doi{10.2106/00004623-198163090-00004}.

\bibitem[Chen et~al.(2025)Chen, Yu, Choi, Cutkosky, and Bohg]{chen2025dexforce}
C.~Chen, Z.~Yu, H.~Choi, M.~Cutkosky, and J.~Bohg.
\newblock Dexforce: Extracting force-informed actions from kinesthetic demonstrations for dexterous manipulation.
\newblock \emph{IEEE Robotics and Automation Letters}, 2025.

\bibitem[Fang et~al.(2025)Fang, Wang, Wang, Chen, Xia, Lv, He, Yi, Guo, Zhan, Yang, Wang, Lu, and Fang]{fang2025airexo}
H.~Fang, C.~Wang, Y.~Wang, J.~Chen, S.~Xia, J.~Lv, Z.~He, X.~Yi, Y.~Guo, X.~Zhan, L.~Yang, W.~Wang, C.~Lu, and H.-S. Fang.
\newblock Airexo-2: Scaling up generalizable robotic imitation learning with low-cost exoskeletons.
\newblock \emph{arXiv preprint arXiv:}, 2025.

\bibitem[An et~al.(1986)An, Askew, and Chao]{an1986biomechanics}
K.~An, L.~Askew, and E.~Chao.
\newblock Biomechanics and functional assessment of upper extremities.
\newblock In \emph{Trends in ergonomics/human factors III}, volume 1986, pages 573--580. Elsevier Science Publishers North-Holland, 1986.

\bibitem[Napier(1956)]{Napier1956}
J.~R. Napier.
\newblock The prehensile movements of the human hand.
\newblock \emph{The Journal of Bone and Joint Surgery. British Volume}, 38-B\penalty0 (4):\penalty0 902--913, 1956.
\newblock \doi{10.1302/0301-620X.38B4.902}.

\bibitem[Marzke et~al.(1992)Marzke, Wullstein, and Viegas]{Marzke1992}
M.~W. Marzke, K.~L. Wullstein, and S.~F. Viegas.
\newblock Evolution of the power (“squeeze”) grip and its morphological correlates in hominids.
\newblock \emph{American Journal of Physical Anthropology}, 89\penalty0 (3):\penalty0 283--298, 1992.
\newblock \doi{10.1002/ajpa.1330890303}.

\bibitem[Heo et~al.(2023)Heo, Lee, Lee, and Lim]{heo2023furniturebench}
M.~Heo, Y.~Lee, D.~Lee, and J.~J. Lim.
\newblock Furniturebench: Reproducible real-world benchmark for long-horizon complex manipulation.
\newblock In \emph{Robotics: Science and Systems}, 2023.

\end{thebibliography}

\clearpage
\section*{Supplementary}
\paragraph{S1. Electronics of \dexop:}
\label{supp:electronics}
Each revolute joint in the passive robotic hand is equipped with an iC-MH16 12-bit angular encoder, offering a resolution of 1.5e-3 radians. An RS-485 interface IC is used to transmit the encoder signals. We designed a custom PCB to aggregate all RS-485 signals and relay them to a host computer via USB. The same PCB also supplies power to the tactile sensors. For the tactile camera modules, we use IMX219 color sensors with fisheye lenses. Data from multiple cameras is collected using Arducam 8MP×4 quadrascopic camera bundle kits.

\paragraph{S2. Design of \dexopn and \dexops:} 
\label{supp:kine}
We illustrate the kinematics of the \dexopn and \dexops in Figure~\ref{fig:kine-supp}. Compared to \dexopt, \dexopn remove the ring finger, and \dexops further remove the abduction at the MCP joints.
\renewcommand{\thefigure}{S\arabic{figure}}

\setcounter{figure}{0}

\begin{figure*}[h]
  \centering
\includegraphics[width=0.8\textwidth]{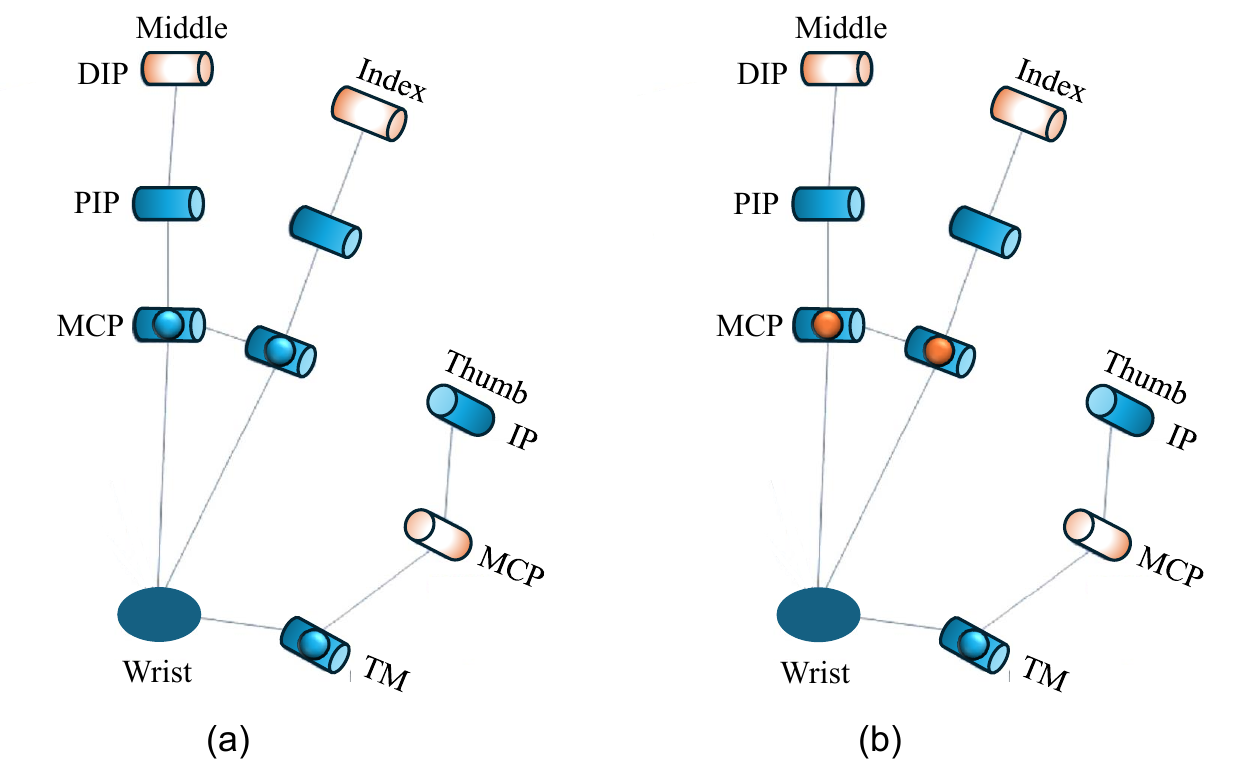}
  \caption{(a) Blue joints represent the kinematic chain of the \dexopn. (b) Blue joints represent the kinematic chain of the \dexops. Orange joints are the missing joints compared to human hand kinematics.}
  \label{fig:kine-supp}
  \vspace{-0.1in}
\end{figure*}

The implementation of \dexopn closely follows the implementation of \dexopn but just removes the ring finger. 

\begin{wrapfigure}{r}{0.4\textwidth} 
  \centering
\vspace{-0.2in}
\includegraphics[width=0.4\textwidth]{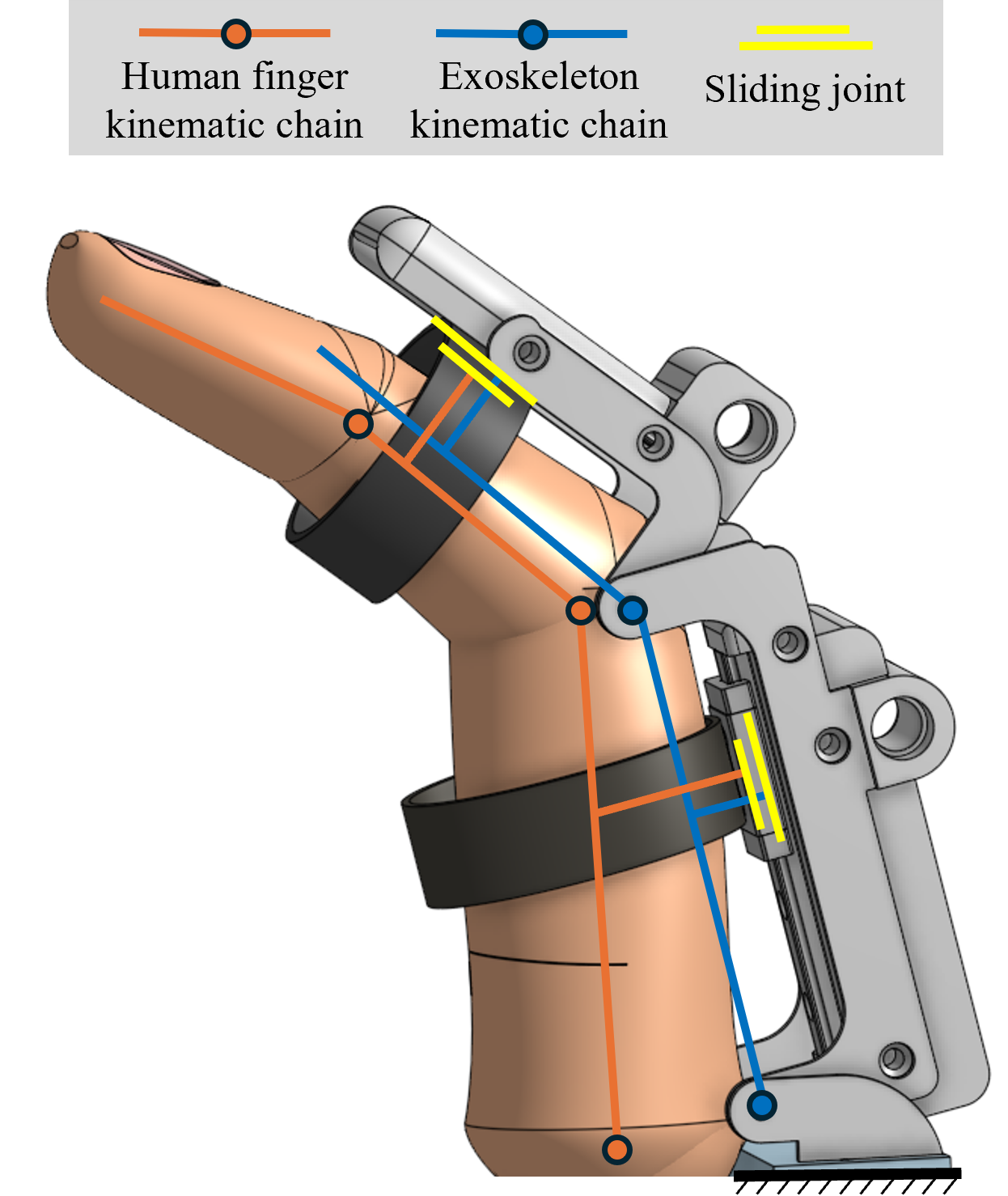}
  \caption{Illustration of attaching distal and proximal phalanges of human finger to wearable exoskeleton on \dexops, which provides force feedback at each finger segment.}
  \label{fig:suppkin}
  \vspace{-0.3in}
\end{wrapfigure}
For the implementation of \dexops, 
besides removing the abduction at the MCP joints, we also explore another method to connect human hand to the wearable exoskeleton, shown in Figure.~\ref{fig:suppkin}. Instead of using a fingertip cot, we attach both the distal and the proximal phalanges of the finger to the exoskeleton using velcro fastener (shown in black). By attaching both phalanges to the exoskeleton, we can provide force feedback to human users on each of those finger segments. However, since the kinematics of the exoskeleton and the human finger are not perfectly aligned, relative sliding between them would occur when the operator bends their fingers. Thus, each finger segment is attached to a linear slider (shown as a sliding joint in the figure), which is fixed on the exoskeleton and can compensate for relative sliding.

\paragraph{S3. Co-design of \dexops and EyeSight Hand:}
\label{supp:codesign}
The development of \dexops was closely coupled with the design of the EyeSight Hand. We began by constructing the passive robotic hand component of \dexops. As discussed in Section~\ref{sec:design}, one of our guiding principles was to match the kinematic chain of the human hand, which required both the exoskeleton and the robotic hand to follow anatomically similar joint configurations.

Fortunately, the EyeSight Hand was originally designed with this goal in mind—it already aligned well with human hand kinematics. Therefore, the initial version of \dexops was built to mirror the existing EyeSight Hand.

However, early testing revealed a critical mismatch: the thumb’s interphalangeal (IP) joint on the EyeSight Hand was designed to bend inward toward the palm to support precise grasping. In contrast, the human IP joint bends outward. This discrepancy made it difficult for users to wear the exoskeleton naturally.

\begin{figure*}[h]
  \centering
\includegraphics[width=0.8\textwidth]{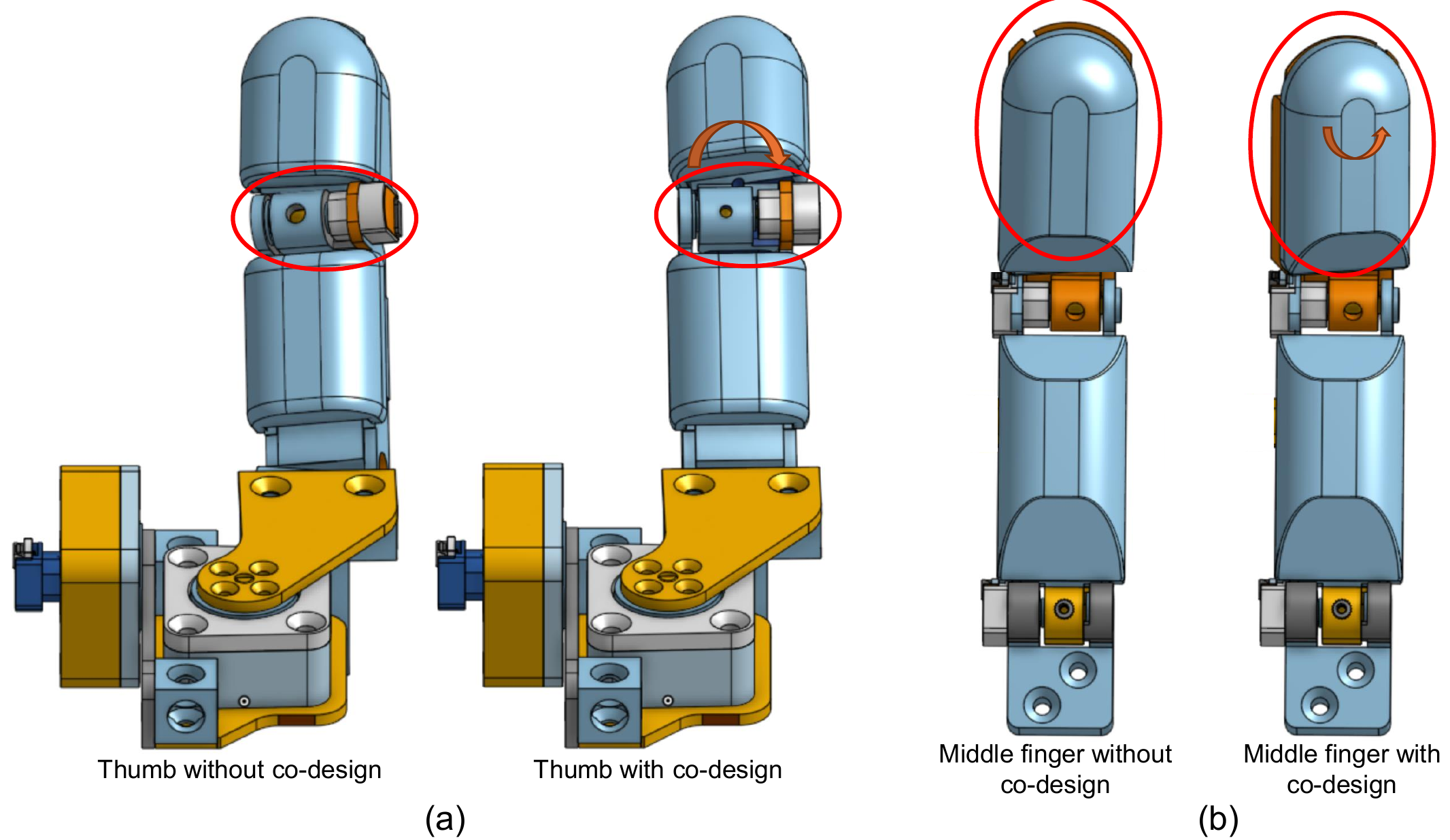}
  \caption{(a) The thumb on passive robotic hand without and with co-designing with the actual robotic hand. (b) The middle finger on the passive robotic hand without and with co-designing with the actual robotic hand.}
  \label{fig:co-design}
  \vspace{-0.1in}
\end{figure*}

To address this, we redesigned the passive robotic hand’s thumb to orient the IP joint outward, consistent with human anatomy (see Figure~\ref{fig:co-design} (a)). But simply modifying the thumb reduced its contact area with the middle fingertip during precise grasping. To compensate, we also adjusted the middle finger by tilting its fingertip (see Figure~\ref{fig:co-design} (b), thereby restoring a wide contact area for thumb-middle finger interactions.

These two key modifications (adjusting the thumb and middle finger) were incorporated back into the EyeSight Hand. This ensures consistency between the exoskeleton and the robotic hand, enabling smoother policy transfer between human demonstrations and robot execution.

\paragraph{S4. Details of hardware enhancements:} 
\label{supp:enhancements}
The design principles discussed in Section~\ref{sec:design} and Section~\ref{subsec:linkage-design} enable perioperation of a passive robotic hand, but simply having a multi-finger hand does not guarantee dexterous capabilities. We observed that the curved fingerpad and bulky backs of the fingers make it difficult to perform antipodal precision grasps on small objects placed on a table, as the backs of the fingers tend to collide with the surface. Reducing finger thickness and adding fingernails enables \dexop to manipulate small objects, such as picking up a coin from a table or retrieving an M2 screw from a pile. In-hand reorientation often requires substantial side contact, so incorporating abduction at the MCP joints and increasing the lateral contact surface of the fingers improves the ability to grip objects between them. The palm is designed to be elastic, allowing it to press objects firmly against the distal phalanges. Finally, the exoskeleton’s finger cots are adjustable, enabling adaptation to different finger lengths and providing more intuitive control.

\begin{figure*}[h]
  \centering
\includegraphics[width=0.8\textwidth]{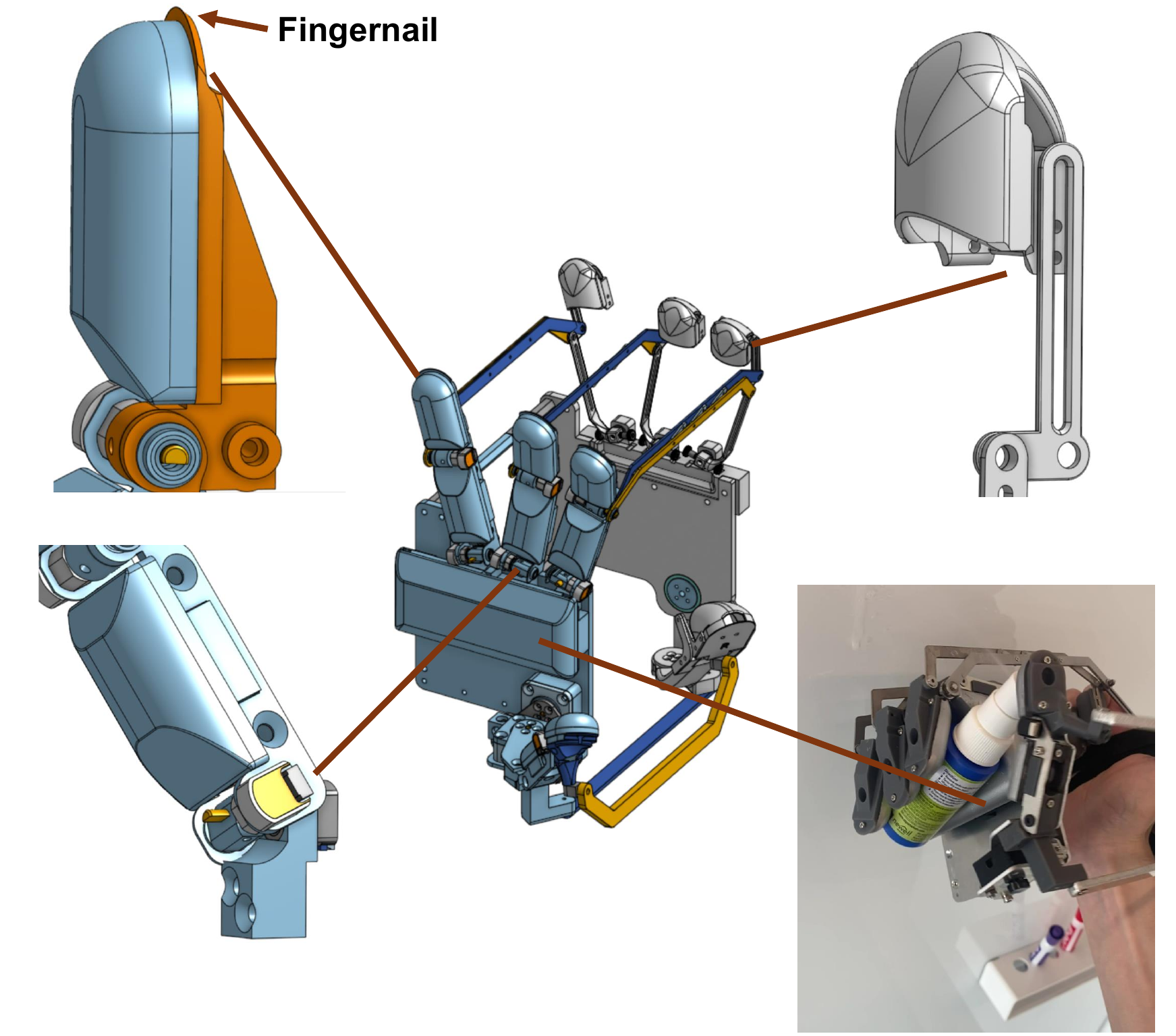}
  \caption{Fingernails (top left) assist in picking up small objects. Abduction joints (bottom left) enable in-hand reorientation and gripping between fingers. Adjustable finger cots (top right) adapt the exoskeleton to different finger lengths. The palm pad (bottom right) secures objects during whole-hand manipulation.}
  \label{fig:enhance}
  \vspace{-0.1in}
\end{figure*}
\end{document}